# Diversity and degrees of freedom in regression ensembles

Henry WJ Reeve    Gavin Brown

*University of Manchester - School of Computer Science*
*Kilburn Building, University of Manchester, Oxford Rd, Manchester M13 9PL*

**Abstract**

Ensemble methods are a cornerstone of modern machine learning. The performance of an ensemble depends crucially upon the level of diversity between its constituent learners. This paper establishes a connection between diversity and degrees of freedom (i.e. the capacity of the model), showing that diversity may be viewed as a form of *inverse regularisation*. This is achieved by focusing on a previously published algorithm *Negative Correlation Learning* (NCL), in which model diversity is explicitly encouraged through a diversity penalty term in the loss function. We provide an exact formula for the effective degrees of freedom in an NCL ensemble with fixed basis functions, showing that it is a continuous, convex and monotonically increasing function of the diversity parameter. We demonstrate a connection to Tikhonov regularisation and show that, with an appropriately chosen diversity parameter, an NCL ensemble can always outperform the unregularised ensemble in the presence of noise. We demonstrate the practical utility of our approach by deriving a method to efficiently tune the diversity parameter. Finally, we use a Monte-Carlo estimator to extend the connection between diversity and degrees of freedom to ensembles of deep neural networks.

*Keywords:* Degrees of freedom, Negative Correlation Learning, Tikhonov regularisation, ensembles, Stein's unbiased risk estimate, deep neural networks.

## 1. Introduction

Ensemble methods are a cornerstone of modern machine learning. Numerous applications have shown that by combining a multiplicity of models we are able to train powerful estimators from large data sets in a tractable way. Successful ensemble performance emanates from a fruitful trade-off between the individual accuracy of the models and their diversity [10]. Typically diversity is introduced implicitly, by sub-sampling the data or varying the architecture of the models. In this paper we consider Negative Correlation Learning (NCL) [32], a powerful approach to learning ensembles of neural networks, in which diversity is encouraged explicitly by appending a diversity penalty term to the loss function. In the context of the recent breakthroughs in deep neural networks, ensembles of neural networks are likely to play an increasingly prominent role in machine learning applications. Thus, it is crucial that we obtain a deeper understanding of the dynamics of ensemble methods well suited to neural networks such as NCL. The statistical properties of NCL have already been studied in some detail [32, 10, 11]. Nonetheless, important questions remain surrounding the *diversity parameter*, the central hyperparameter in NCL which controls the level of emphasis placed upon the diversity penalty term. We shall address the following:

- How does the complexity of the ensemble estimator vary as a function of the diversity parameter?
- How can we efficiently optimise the diversity parameter on large data sets?
- Is the optimal value of the diversity parameter always strictly less than one?

The core of our investigation lies in a degrees of freedom analysis of NCL ensembles. Our contributions are as follows:

- We derive a formula for the degrees of freedom under the assumption of fixed basis functions (Section 3).
- We show analytically that the degrees of freedom is monotonically increasing as a function of the diversity parameter (Section 3).

- We present the surprising result that, in the presence of noise, the optimal value of the diversity parameter is always strictly less than one (Section 4).

- We develop an intriguing connection between NCL and Tikhonov regularisation (Section 5).

- We present an empirical verification of the theoretical results (Section 6).

- We give a fast and effective procedure for tuning the diversity parameter based upon the degrees of freedom (Section 7).

- We investigate ensembles of deep neural networks, demonstrating empirically that the degrees of freedom also behave monotonically with respect to the diversity parameter in this setting (Section 8).

The present paper extends a previously published conference paper [40]. The previous conference paper introduces the analytic formula for the degrees of freedom and demonstrates a computationally efficient approach to tuning the diversity parameter based on the formula. In the present paper we have extended this work. Firstly, we present additional technical results: A connection between NCL and Tikhonov regularisation and a result implying that the diversity parameter should never be set to precisely one in the presence of noise. Secondly, we used a Monte-Carlo estimator to conduct a detailed empirical investigation into the relationship between the diversity parameter and degrees of freedom in ensembles of deep neural networks.

We shall begin by introducing the background on ensemble learning and degrees of freedom in Section 2.

## 2. Background

In this section we shall introduce the relevant background on Negative Correlation Learning (NCL) and degrees of freedom. We begin by setting the scene. Throughout this paper we consider the regression problem: We are given a data set $\mathcal{D} = \{(x_n, y_n)\}_{n=1}^N$ with $(x_n, y_n) \in \mathcal{X} \times \mathbb{R}$. We shall assume that there is an underlying function $\mu : \mathcal{X} \to \mathbb{R}$ such that for each $n$, $y_n = \mu(x_n) + \epsilon_n$, where $(\epsilon_n)_{n=1}^N$ is a mean zero, independent and identically distributed random process. Our goal is to use the data $\mathcal{D}$ to provide an estimator $\hat{\mu} : \mathcal{X} \to \mathbb{R}$ of the underlying function $\mu$.

*2.1. Ensembles, diversity and the ambiguity decomposition*

Ensemble methods aggregate the predictions of a multiplicity of constituent models in order to provide a more powerful model with lower generalisation error. In order for an ensemble to outperform a single model it is essential for its constituent models to be *diverse* [8]. In the classification setting, there is no straightforward relationship between the performance of an ensemble and its diversity [17]; the ensemble error can even exceed the average error of its constituent learners. In the regression setting, however, the squared error of the ensemble may be decomposed into the average squared error of its constituents minus the variance over the ensemble's predictions. To be precise, suppose we have an ensemble $\mathcal{F} = \{f_m\}_{m=1}^M$ consisting of $M$ functions $f_m : \mathcal{X} \to \mathbb{R}$. We let $F := (1/M) \cdot \sum_{m=1}^M f_m$ denote the ensemble function. For each $(x, y) \in \mathcal{X} \times \mathbb{R}$ we have,

$$\overbrace{(F(x) - y)^2}^{\text{ensemble error}} = \overbrace{\frac{1}{M} \sum_{m=1}^M (f_m(x) - y)^2}^{\text{average error}} - \overbrace{\frac{1}{M} \sum_{m=1}^M (f_m(x) - F(x))^2}^{\text{diversity}}. \tag{1}$$

This relationship is known as the *ambiguity decomposition*. It was observed by Krogh and Vedelsby who highlighted its importance for ensemble learning [28]. We refer to the variance over the ensemble's outputs as the *diversity*. The ambiguity decomposition shows that the square error of the ensemble never exceeds the average error of its constituent learners, and the extent to which the ensemble outperforms its constituents is quantified by its diversity.

Hence, ensemble methods succeed by attaining a high degree of diversity without sacrificing too much individual accuracy. Typically ensemble methods encourage diversity implicitly by modifying the training data or model-structure for the constituent models. For example, Ada-boost encourages diversity by increasing the weight of examples mis-classified by previous models [15], whereas random forests encourage diversity by training different trees with different bootstrap samples of the data and splitting branches along different subsets of features. However, the ambiguity decomposition (1) motivates a more direct approach: *Negative Correlation Learning*.

*2.2. Negative Correlation Learning*

The Negative Correlation Learning method, introduced by Liu and Yao [32], encourages diversity explicitly by incorporating a diversity penalty term into the cost function. Suppose we have an ensemble $\mathcal{F} = \{f_m\}_{m=1}^{M}$ with each function $f_m : \mathcal{X} \to \mathbb{R}$ chosen from a parameterisable family of neural networks $\mathcal{H}_m$, parameterised by $\theta_m$. Our ensemble estimator $\hat{\mu}$ is given by the average $F = (1/M) \cdot \sum_{m=1}^{M} f_m$. The NCL rule, introduced by Liu and Yao [32] proceeds as follows. First each parameter vector $\theta_m$ is randomly initialised. Then, for each training example $(x_n, y_n) \in \mathcal{D}$, we update each $\theta_m$ in parallel according to

$$\theta_m \leftarrow \theta_m - \alpha \cdot \frac{\partial f_m}{\partial \theta_m} \cdot \left( \overbrace{(f_m(x_n) - y)}^{\text{accuracy}} - \lambda \cdot \overbrace{(f_m(x_n) - F(x_n))}^{\text{diversity}} \right),$$

where $\alpha$ is a learning rate. Thus, each update consists of two components: The first pushes the output of the model in the direction of the target, making the model more accurate over the training data. The second pushes the individual model output away from average output, encouraging diversity. The NCL rule is equivalent to stochastic gradient descent with respect to the following loss function (with a scaled learning rate),

$$L_\lambda(\mathcal{F}, x, y) := \overbrace{\frac{1}{M} \sum_{m=1}^{M} (f_m(x) - y)^2}^{\text{accuracy}} - \lambda \cdot \overbrace{\frac{1}{M} \sum_{m=1}^{M} (f_m(x) - F(x))^2}^{\text{diversity}}. \quad (2)$$

We shall refer to $L_\lambda$ as the NCL loss.

The study of NCL is important for several reasons. Firstly the NCL method has been shown to perform well on a wide variety of regression problems, in some cases significantly outperforming other ensemble methods such as boosting and bagging [32, 10]. Secondly, NCL holds a privileged place amongst ensemble methods due to its explicit emphasis upon diversity. Thirdly, the past decade has seen phenomenal progress in deep learning with artificial neural networks surpassing human performance on certain tasks [4, 27, 30, 41, 18]. NCL is specifically designed for generating ensembles of neural networks. Hence, there is a great potential for future applications of NCL to deep neural networks.

The key focus of this paper will be understanding the behaviour of the ensemble as a function of the diversity parameter $\lambda$, which explicitly manages a trade-off between the two competing objectives of accuracy and diversity. An important observation of Brown et al. is that the NCL loss may be rewritten in the following way [9, 10]:

$$L_\lambda(\mathcal{F}, x, y) := (1 - \lambda) \cdot \overbrace{\frac{1}{M} \sum_{m=1}^{M} (f_m(x) - y)^2}^{\text{individual accuracy}} + \lambda \cdot \overbrace{(F(x) - y)^2}^{\text{combined accuracy}}. \quad (3)$$

Hence, when $\lambda = 0$ each function $f_m$ is trained individually and when $\lambda = 1$, $L_\lambda$ is the squared error for the average $F$. Hence, NCL scales smoothly between training each of the functions $f_m$ individually and training as a single combined estimator $F$. Brown et al. conducted a detailed analysis of NCL, relating the behaviour of the ensemble to the bias-variance-covariance decomposition [10]. In addition, Brown et al. gave an upper bound on the diversity parameter $\lambda$, showing that for $\lambda > M/(M-1) > 1$ the Hessian matrix of the weights is non-positive semi-definite. It was subsequently shown that for any $\lambda > 1$, minimising $L_\lambda$ causes the weights to diverge [39, Theorem 3]. Thus, we should restrict the diversity parameter $\lambda$ to the region $\lambda \in [0, 1]$. Nonetheless, many open questions remain.

*2.3. Open questions in Negative Correlation Learning*

Figure 1 displays the typical behaviour of the mean squared error as a function of the diversity parameter $\lambda$. As the diversity parameter moves from zero to one, the training error declines. This is to be expected given that $L_\lambda$ with $\lambda = 1$ corresponds to the square error for the ensemble $F$. The test error typically declines initially before rising sharply as $\lambda$ approaches one. Whilst the test error does not always rise sharply as $\lambda \to 1$ (see the Kinematics data set in Figure 1), we do consistently observe an increase in the gap between test and train error (we shall discuss the Kinematics data set in more detail in Section 4). It is this phenomenon that we intend to explain. Note that this increase cannot be explained by the upper bounds in [10] and [39], since these relate to behaviour on the training set $\mathcal{D}$ for $\lambda > 1$.

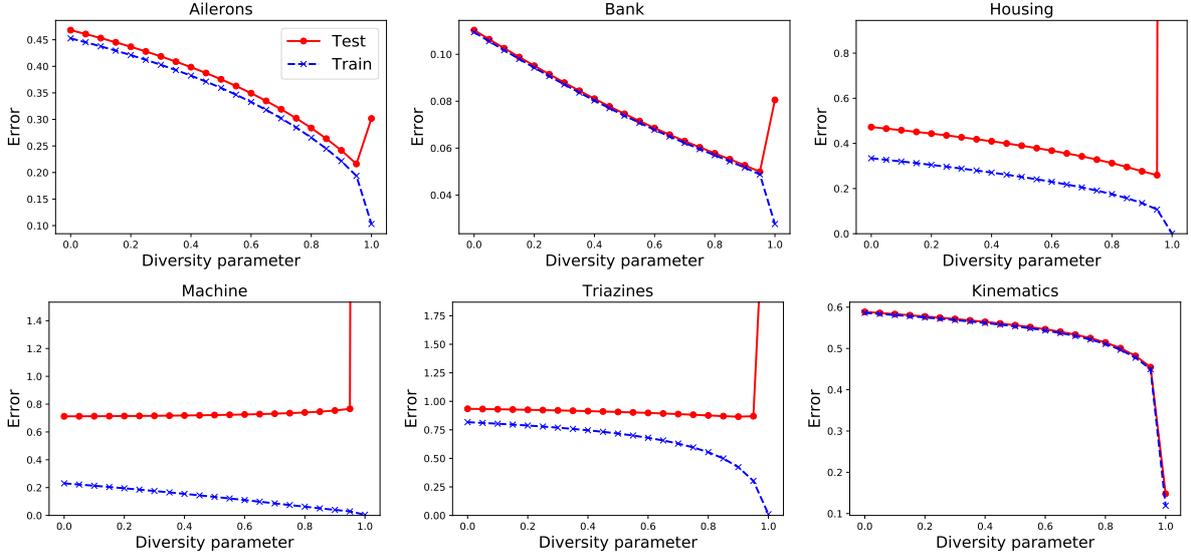

Figure 1: The train and test (mean squared) error for Negative Correlation Learning ensembles $F_\lambda$ as a function of the diversity parameter. In each case we consider an ensemble of $M = 100$ models with $H = 10$ hidden nodes per model $f_m$. For precise experimental details see Section 6.

Taking $\lambda < 1$ appears to act as a regulariser, reducing the discrepancy between test and train error. This is intuitively plausible, given equation 3: When $\lambda = 0$ we are independently training a collection of $M$ simple models and aggregating the result, an approach which is likely to under-fit. When $\lambda = 1$ we are minimising the training error for a single complex model $F$, which is likely to overfit. Choosing $\lambda$ between zero and one blends smoothly between these extremes, providing an effective balance between underfitting and overfitting. Nonetheless, the underlying hypothesis class does not change with $\lambda$. Hence, the usual formalisms of VC dimension or Rademacher complexity do not apply. This raises several questions. Firstly, how can we provide a rigorous explanation for the apparent regularisation effect of taking the diversity parameter $\lambda < 1$, in spite of the fact that the hypothesis class remains unchanged. Secondly, under what circumstances does taking $\lambda < 1$ improve performance? Finally, how can we optimise the diversity parameter on large data sets without resorting to an expensive cross-validation procedure. To address these questions we shall use the concept of *degrees of freedom*.

### 2.4. Degrees of freedom and Stein's unbiased risk estimate

In this section we shall introduce the relevant background on degrees of freedom. We shall operate within the the fixed design setting, frequently utilised in statistics [14, 50, 44]. To be precise, suppose we have a fixed design matrix $X = [x_1, \cdots, x_N]$ consisting of feature vectors $x_n \in \mathcal{X}$ and assume the existence of an underlying function $\mu : \mathcal{X} \to \mathbb{R}$. Our data set $\mathcal{D} = \{(x_n, y_n)\}_{n=1}^{N}$ consists of pairs $(x_n, y_n) \in \mathcal{X} \times \mathbb{R}$ generated by $y_n = \mu(x_n) + \epsilon_n$. Here $(\epsilon_n)_{n=1}^{N}$ is a mean zero, independent and identically distributed (i.i.d) random process with variance $\sigma^2 = \mathbb{E}\left[\epsilon_n^2\right]$. Given an estimator $\hat{\mu}$ for $\mu$ we define the true error $R_{\text{true}}$ and the empirical error $R_{\text{emp}}$, respectively, by

$$R_{\text{true}}(\hat{\mu}) = \frac{1}{N} \sum_{n=1}^{N} (\hat{\mu}(x_n) - \mu(x_n))^2 \qquad R_{\text{emp}}(\hat{\mu}, \mathcal{D}) = \frac{1}{N} \sum_{n=1}^{N} (\hat{\mu}(x_n) - y_n)^2 .$$

The key tool in our analysis of NCL will be the degrees of freedom.

**Definition 1** (Degrees of freedom). *Let $\hat{\mu}$ be an estimator for $\mu$. The degrees of freedom of $\hat{\mu}$ is defined by*

$$df(\hat{\mu}) := \sum_{n=1}^{N} \frac{\partial \hat{\mu}(x_n)}{\partial y_n}.$$

In the ordinary least squares setting $df(\hat{\mu})$ is simply the number of non-trivial parameters in the model, hence the nomenclature. However, in general the degrees of freedom is not determined by the number of model parameters [48]. The degrees of freedom quantifies the complexity of an estimation procedure via the sensitivity of model outputs to the data. A fundamental result of Stein implies that the degrees of freedom determines the relationship between the empirical error and the true error of an estimator.

**Definition 2** (Stein's unbiased risk estimate). *Given an estimator $\hat{\mu}$, a data set $\mathcal{D}$, and a positive real $\tilde{\sigma} > 0$, Stein's unbiased risk estimate is defined by*

$$SURE(\hat{\mu}, \mathcal{D}, \tilde{\sigma}) := R_{emp}(\hat{\mu}, \mathcal{D}) + \tilde{\sigma}^2 \cdot \left( \frac{2}{N} \cdot df(\hat{\mu}, \mathcal{D}) - 1 \right).$$

**Theorem 3** (Stein [43]). *Suppose that the noise process $(\epsilon_n)_{n=1}^N$ is i.i.d with $\epsilon_n \sim \mathcal{N}(0, \sigma^2)$. Given any estimator $\hat{\mu}$ such that for each $n = 1, \cdots, N$, $\hat{\mu}(x_n)$ is differentiable with respect to $y_n$, we have*

$$\mathbb{E}\left[R_{true}(\hat{\mu})\right] = \mathbb{E}\left[SURE(\hat{\mu}, \mathcal{D}, \sigma)\right].$$

Theorem 3 implies that the true error may be broken down, in expectation, into two concrete factors: The empirical error and the degrees of freedom. In Section 3 we shall use this decomposition to give a theoretical explanation for the empirical behaviour of NCL ensembles described in Section 2. Theorem 3 also has important practical implications for model selection. Suppose we have a family of estimators $\mathcal{A}$. We would like to choose the estimator $\hat{\mu} \in \mathcal{A}$ which optimises performance by minimising $R_{\text{true}}(\hat{\mu})$ over $\mathcal{A}$. Unlike the true error $R_{\text{true}}(\hat{\mu})$, the empirical error $R_{\text{emp}}(\hat{\mu}, \mathcal{D})$ may be computed directly from the data set. However, $R_{\text{emp}}(\hat{\mu}, \mathcal{D})$ is a biased estimate for the true error $R_{\text{true}}(\hat{\mu})$ with the bias depending upon the sensitivity of the estimator. Hence, if we minimise $R_{\text{emp}}(\hat{\mu}, \mathcal{D})$ over $\hat{\mu} \in \mathcal{A}$ we are likely to obtain a highly complex estimator which overfits to the training data. Theorem 3 states that $SURE(\hat{\mu}, \mathcal{D}, \sigma)$ gives an *unbiased* estimate for the true error $R_{\text{true}}(\hat{\mu})$. This motivates a strategy where we select $\hat{\mu} \in \mathcal{A}$ by minimising $SURE(\hat{\mu}, \mathcal{D}, \tilde{\sigma})$, provided we are able to compute $df(\hat{\mu}, \mathcal{D})$ and obtain an estimate $\tilde{\sigma}$ for the noise variance $\sigma$. In Section 7 we use this approach to give a highly efficient procedure for tuning the diversity parameter in NCL ensembles.

## 3. A theoretical investigation of the degrees of freedom in Negative Correlation Learning

In this section we give a theoretical analysis of the degrees of freedom of NCL ensembles. We consider a special class of NCL ensembles $\mathcal{F}$ in which each function $f_m$ consists of a linear map applied to a fixed basis function $\phi_m$. More precisely, for each $m = 1, \cdots, M$ we take a basis function $\phi_m : \mathcal{X} \to \mathbb{R}^{H \times 1}$ which does not depend upon the targets $\{y_n\}$ but may depend upon the design matrix $X$. We let $\mathcal{H}_m = \left\{ x \mapsto \langle w_m, \phi_m(x) \rangle : w_m \in \mathbb{R}^{H \times 1} \right\}$, where $\langle \cdot, \cdot \rangle$ denotes the dot product, and assume that $f_m \in \mathcal{H}_m$ (see Figure 2). This encompasses a broad range of highly expressive mappings depending upon the choice of basis function. Examples include Nyström features [47, 12] and Random Fourier Features [36, 37] (see Section 6 for details). We consider ensembles $\mathcal{F} = \{f_m\}_{m=1}^M$ where each $f_m \in \mathcal{H}_m$. The assumption of fixed basis functions allows us to analytically derive several properties of the ensemble that we would otherwise only be able to observe empirically.

The assumption of fixed basis functions allows us to derive a closed form solution for the minimiser of $L_\lambda$. We first introduce some notation. Let $Q = H \cdot M$ and let $\boldsymbol{\phi} : \mathcal{X} \to \mathbb{R}^{Q \times 1}$ be the function defined by $\boldsymbol{\phi}(x) = [\phi_1(x)^T, \cdots, \phi_M(x)^T]^T$. We define

$$\langle \boldsymbol{\phi}, \boldsymbol{\phi} \rangle_\mathcal{D} := \frac{1}{N} \sum_{n=1}^N \boldsymbol{\phi}(x_n) \boldsymbol{\phi}(x_n)^T \in \mathbb{R}^{Q \times Q}$$

$$\langle \boldsymbol{\phi}, y \rangle_\mathcal{D} := \frac{1}{N} \sum_{n=1}^N y_n \boldsymbol{\phi}(x_n) \in \mathbb{R}^{Q \times 1}$$

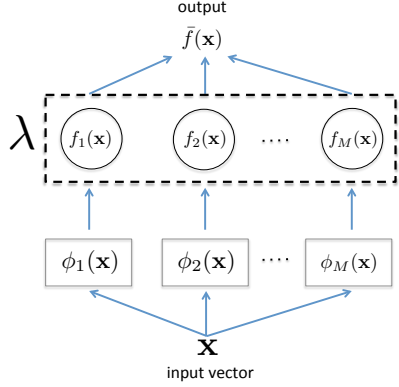

Figure 2: Illustration of the fixed basis function scenario for Negative Correlation Learning.

Similarly, for each $m = 1, \cdots, M$, we define

$$\langle \phi_m, \phi_m \rangle_{\mathcal{D}} := \frac{1}{N} \sum_{n=1}^{N} \phi_m(x_n) \phi_m(x_n)^T \in \mathbb{R}^{H \times H}$$

$$\langle \boldsymbol{\phi}, \boldsymbol{\phi} \rangle_{\mathcal{D}}^{\text{diag}} := \text{diag}\left( \langle \phi_1, \phi_1 \rangle_{\mathcal{D}}, \cdots, \langle \phi_M, \phi_M \rangle_{\mathcal{D}} \right) \in \mathbb{R}^{Q \times Q}.$$

We shall assume that $\langle \phi_m, \phi_m \rangle_{\mathcal{D}}$ is of full rank $H$. This assumption is natural, when $H \ll N$, as is the case in typical applications of NCL [10, 32]. The assumption only fails to hold when one of the $H$ coordinates of a basis function $\phi_m$ is expressible as a linear combination of the other $H-1$ coordinates of $\phi_m$, over all $N$ examples. This is a strong form of redundancy which we are unlikely to encounter in practice.

Given a matrix $M$ we let $M^+$ denote the Moore-Penrose psuedo-inverse of $M$. Given $\lambda \in [0, 1]$, we let $F_\lambda$ denote the ensemble function $F_\lambda = (1/M) \cdot \sum_{m=1}^{M} f_m^\lambda$ where $\mathcal{F}_\lambda = \{f_m^\lambda\}_{m=1}^{M}$ denotes the ensemble which minimises the NCL loss $L_\lambda$ averaged over a data set, i.e.

$$\mathcal{F}_\lambda = \text{argmin} \left\{ \frac{1}{N} \sum_{n=1}^{N} L_\lambda(\mathcal{F}, x_n, y_n) : \mathcal{F} = \{f_m\}_{m=1}^{M}, f_m \in \mathcal{H}_m \right\}.$$

**Theorem 4.** *Suppose that each function $f_m$ consists of a linear map applied to a fixed basis function $\phi_m$. For each $\lambda \in [0, 1]$, the ensemble function $F_\lambda : \mathcal{X} \to \mathbb{R}$ is given by $F_\lambda(x) = \langle \beta_\lambda, \boldsymbol{\phi}(x) \rangle$ where*

$$\beta_\lambda = \left( M(1-\lambda) \cdot \langle \boldsymbol{\phi}, \boldsymbol{\phi} \rangle_{\mathcal{D}}^{\text{diag}} + \lambda \langle \boldsymbol{\phi}, \boldsymbol{\phi} \rangle_{\mathcal{D}} \right)^+ \langle \boldsymbol{\phi}, \mathbf{y} \rangle_{\mathcal{D}}.$$

The closed form solution allows us to efficiently locate the minimiser of the NCL loss. Moreover, Theorem 4 leads to the following theorem which gives an explicit formula for the degrees of freedom of $F_\lambda$, and describes the behaviour of $\text{df}(F_\lambda, \mathcal{D})$ as a function of $\lambda$.

**Theorem 5.** *Suppose that each function $f_m$ is a linear map of a fixed basis function. Given any $\lambda \in [0, 1]$, the degrees of freedom for $F_\lambda$ is given by*

$$\text{df}(F_\lambda, \mathcal{D}) = \text{trace}\left( \langle \boldsymbol{\phi}, \boldsymbol{\phi} \rangle_{\mathcal{D}} \left( M(1-\lambda) \cdot \langle \boldsymbol{\phi}, \boldsymbol{\phi} \rangle_{\mathcal{D}}^{\text{diag}} + \lambda \langle \boldsymbol{\phi}, \boldsymbol{\phi} \rangle_{\mathcal{D}} \right)^+ \right).$$

*In particular, we have the following behaviour*

- *The function $\lambda \mapsto \text{df}(F_\lambda, \mathcal{D})$ is continuous, increasing and convex;*
- *The function $\lambda \mapsto R_{\text{emp}}(F_\lambda, \mathcal{D})$ is continuous and decreasing.*

*Moreover, if $H < \text{rank}(\langle \boldsymbol{\phi}, \boldsymbol{\phi} \rangle_\mathcal{D})$ then,*

- *The function $\lambda \mapsto df(F_\lambda, \mathcal{D})$ is strictly increasing and strictly convex;*
- *The function $\lambda \mapsto R_{emp}(F_\lambda, \mathcal{D})$ is strictly decreasing.*

Full proofs for all theorems presented in this paper may be found in the appendix. Theorem 5 is the core result in our analysis. It gives a rigorous account of the empirical behaviour of NCL ensembles observed in Figure 1: As the diversity parameter increases, the degrees of freedom increase monotonically, and at an increasing rate, leading to an increasing discrepancy between the test and train error. This behaviour leads to our view of diversity as a form of *inverse regularisation*. Typically increasing a regularisation parameter leads to more stable models with lower degrees of freedom [50]. Conversely, increasing the diversity parameter produces a less stable model with more degrees of freedom. We shall confirm these results empirically in Section 6.

## 4. Too much diversity?

In this section we answer a fundamental question raised by Brown et al. [10]: Under what conditions does taking the diversity parameter $\lambda < 1$ improve the performance of the NCL ensemble? We shall return to the fixed basis function setting introduced in Section 3 in which each function $f_m$ consists of a linear map applied to a fixed basis function. In this setting, we provide the following surprising answer: Whenever the noise variance $\sigma$ is strictly positive, taking $\lambda = 1$ is always sub-optimal. That is, whenever $\sigma > 0$, there exists a value $\lambda_{\text{opt}} < 1$ such that $R_{\text{true}}(F_{\lambda_{\text{opt}}}) < R_{\text{true}}(F_1)$.

**Theorem 6.** *Suppose that each function $f_m$ is a linear map of a fixed basis function and let $\sigma^2 = \mathbb{E}[\epsilon_n^2]$, the noise variance. Then $1 \in \text{argmin}\{\mathbb{E}[R_{true}(F_\lambda)] : \lambda \in [0, 1]\}$ if and only if $\sigma = 0$.*

In light of equation 3 this result is perhaps surprising - taking $\lambda < 1$ moves the NCL loss $L_\lambda$ away from directly targeting the squared error of the ensemble. However, by Theorem 5 we see that reducing the value of the diversity parameter also reduces the degrees of freedom, making the ensemble more robust to noise in the data. Theorem 6 goes further, showing that whenever the noise is positive there exists a $\lambda < 1$ such that the reduction in error due to the fall in degrees of freedom more than compensates for the increase in the training error.

Let's return briefly to the results displayed in Figure 1, Section 2. For the first five data sets we observed a sharp rise in test error when $\lambda = 1$, which conforms neatly to the conclusions of Theorem 6. However, for the Kinematics data set we see that the optimal value of $\lambda \in \{0.00, 0.05, \cdots, 0.95, 1.00\}$ is actually $\lambda = 1.00$. This seems in tension with Theorem 6. There are two possibilities: Either we have a data set where $\sigma = 0$, which seems highly unlikely, or the optimal value of $\lambda$ lies strictly between 0.95 and 1.00. Hence, we repeat the experiment with $\lambda$ close to one: $\lambda \in \{1 - 10^{-l} : l \in \{2, \cdots, 20\}\}$. The results are displayed in Figure 3. As we can see the optimal value of the diversity parameter is $\lambda \approx 1 - 10^{-12} < 1$, which resolves the apparent tension with Theorem 6.

Theorem 6 is curiously reminiscent of a classical result of Hoerl and Kennard on Tikhonov regularisation [23]. Hoerl and Kannard's result states that in the presence of noise there always exists a strictly positive value of the regularisation parameter which outperforms the unregularised least squares estimator. This strengthens our view of diversity as a form of inverse regularisation in which taking the diversity parameter to be strictly less than one bears an interesting parallel to positive Tikhonov regularisation. In the next section we shall provide a more detailed exploration of the connections between NCL and Tikhonov regularisation.

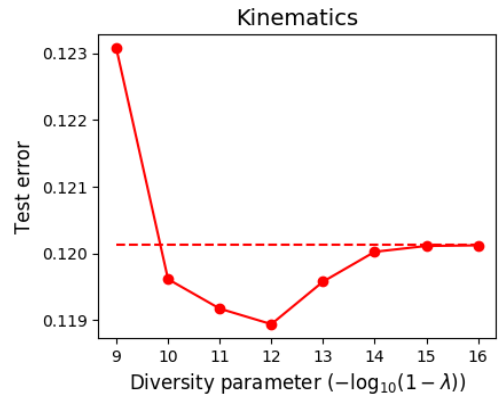

Figure 3: The test error for an NCL ensemble consisting of $M = 100$ models with $H = 10$ hidden nodes per model $f_m$ on the Kinematics data set. We sample the diversity parameter $\lambda$ close to one: $\lambda \in \{1 - 10^{-l} : l \in \{2, \cdots, 20\}\}$. The average test error for $\lambda = 1.0$ is given by the dashed line. For precise experimental details see Section 6.

## 5. Negative Correlation Learning and Tikhonov regularisation

In this section we explore the relationship between NCL ensembles $F_\lambda$ and Tikhonov regularisation. Tikhonov regularisation, also known as weight decay or L2 regularisation, is perhaps the most widely used form of regularisation in machine learning - one simply adds a constant multiple of the *L2* norm of the weights into the cost function. This encourages the weights to be close to the origin with respect to the Euclidean distance, increasing the robustness of the resultant model.

We shall consider the fixed basis function setting in which we have a function of the form $G = (1/M) \cdot \sum_{m=1}^{M} g_m$ where each function $g_m$ consists of a linear weight composed with a fixed basis function, $g_m = \langle w_m, \psi_m \rangle$. Let $\mathcal{G} = \{g_m\}_{m=1}^{M}$. We define the following Tikhonov regularised loss function,

$$L_\gamma^{\text{Tik}}(\mathcal{G}, x, y) = (G(x) - y)^2 + \gamma \cdot \sum_{m=1}^{M} \|w_m\|_2^2.$$

We let $G_\gamma^{\text{Tik}}$ denote the minimiser of the average Tikhonov regularised loss funciton $L_\gamma^{\text{Tik}}$,

$$G_\lambda^{\text{Tik}} = \text{argmin}\left\{\frac{1}{N}\sum_{n=1}^{N} L_\gamma^{\text{Tik}}(\mathcal{G}, \boldsymbol{x}_n, y_n) : \mathcal{G} = \{g_m\}_{m=1}^{M}, g_m = \langle w_m, \psi_m \rangle\right\}.$$

Recall from Section 3 that $F_\lambda$ denotes the function obtained by minimising the NCL loss $L_\lambda$ introduced in equation (3). We have the following relationship between $G_\gamma^{\text{Tik}}$ and $F_\lambda$.

**Theorem 7.** *Suppose that we have a NCL ensemble $F_\lambda = (1/M) \cdot \sum_{m=1}^{M} f_m$ where each function $f_m$ is a linear map of fixed basis function $\phi_m$. Let $G_\gamma^{Tik}$ be the function obtained by Tikhonov regularisation with $G_\gamma^{Tik} = (1/M) \cdot \sum_{m=1}^{M} g_m$, where each map $g_m$ is a linear map applied to a fixed basis function $\psi_m$. Suppose that for each m, $\psi_m(x) = \langle \phi_m, \phi_m \rangle_{\mathcal{D}}^{-1/2} \cdot \phi_m(x)$. Then for all $\lambda \in [0, 1]$, with $\gamma = (1 - \lambda)/(M^2 \cdot \lambda)$ we have $G_\gamma^{Tik} = \lambda \cdot F_\lambda$.*

Theorem 7 gives a close relationship between NCL and Tikhonov regularisation. This connection is especially interesting in light of the range of perspectives on Tikhonov regularisation present in the literature, from Bayesian perspectives [6] through to viewing Tikhonov regularisation as training in the presence of noise [5]. Note that the level of Tikhonov regularisation decays monotonically in relation to the corresponding value of the diversity parameter in Theorem 7. This conforms to the view of diversity as a form of inverse regularisation.

## 6. An empirical investigation of the degrees of freedom curve

In this section we empirically verify the key results in Theorem 5 concerning the behaviour of both the degrees of freedom df $(F_\lambda, \mathcal{D})$ and the training error $R_{\text{emp}}(F_\lambda, \mathcal{D})$ as a function of the diversity parameter $\lambda$. We consider six data sets described in Appendix E. In each case the data was preprocessed by renormalising the targets to mean zero and standard devation one. For each data set we consider three architectures with with five, ten and twenty hidden nodes per module, respectively ie. $H \in \{5, 10, 20\}$. In each case the total number of hidden nodes is set at one thousand ($H \cdot M = 1000$), so the number of modules $M \in \{200, 100, 50\}$.

For each $m = 1, \cdots, M$ we take as our basis function $\phi_m : X \to \mathbb{R}^H$, an $H$-tuple of Random Fourier Features. Random Fourier Features were introduced by Rahimi and Recht [36] and build upon the longstanding tradition of generating the lower layers of a neural network by randomisation, rather than optimisation [7]. These basis functions are of the form $\phi_m(x) = \cos\left(\zeta_m^T x + b_m\right)$, for $x \in \mathbb{R}^{d \times 1}$ in the original feature space, $\zeta_m \in \mathbb{R}^{H \times d}$, $b_m \in \mathbb{R}^{H \times 1}$ and cos applied elementwise. Rahimi and Recht proved that, given any stationary Mercer kernel $k(x, y) = k(x - y)$ on $\mathbb{R}^d$, if we sample the rows of $\zeta_m$ randomly from the normalised Fourier transform of $k$, and sample the elements of $b_m$ uniformly from $[0, 2\pi)$, then we have $(1/H) \cdot \langle \phi_m(x_\alpha), \phi_m(x_\beta) \rangle \approx k(x_\alpha, x_\beta)$, with an approximation error decaying exponentially with $H$. Thus, we may view Random Fourier Features $\phi_m$ constructed in this way as the composition of two mappings: A mapping to the reproducing kernel Hilbert space, composed with a low-dimensional projection which approximately preserves the inner-product structure. In our experiments we use the Scikit-Learn implementation [35], with the Gaussian kernel, which is its own Fourier transform (up to rescaling), so rows of $\zeta_m$ are sampled from a $d$-dimensional Gaussian. We use these basis functions as they are highly expressive and have been successfully used in a range of applications [36, 37].

In order to tune the frequency parameter $\gamma$ for the random Fourier functions using the heuristic method of Kwok and Tsang [29]. For each $\lambda \in \{0, 0.5, \cdots, 0.95, 1.0\}$ we train an NCL estimator $\boldsymbol{F}_\lambda$. The weights $\boldsymbol{w}_m$ are selected using the analytic solution given in Theorem 4, rather than via stochastic gradient descent. Note that this has the benefit of removing the need to select a learning rate.

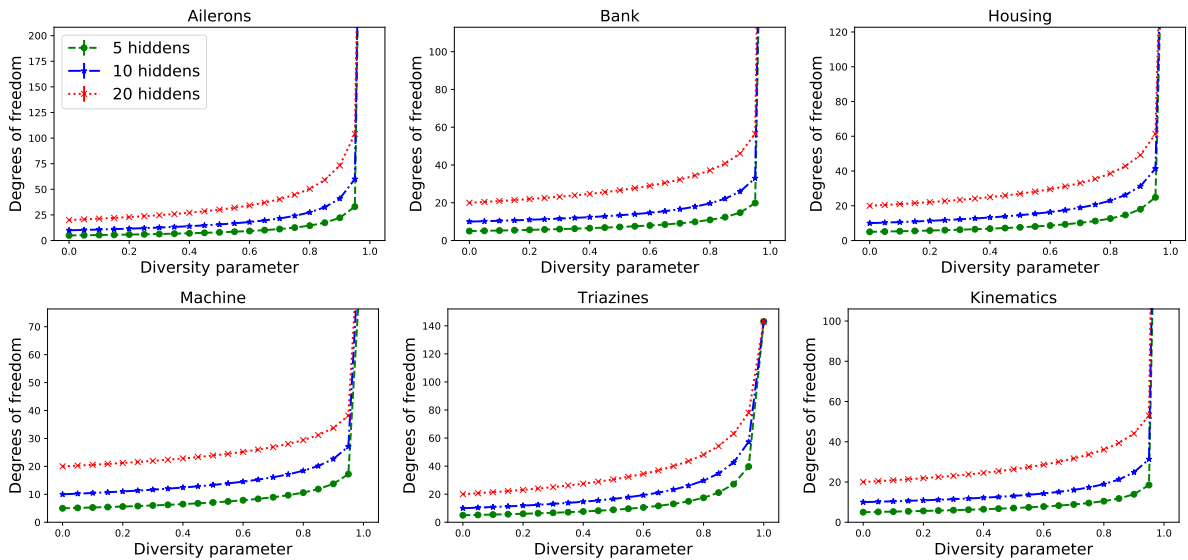

Figure 4: The degrees of freedom df $(\boldsymbol{F}_\lambda, \mathcal{D})$ as a function of the diversity parameter $\lambda$. For details see section 6.

Figure 4 shows the degrees of freedom df $(\boldsymbol{F}_\lambda, \mathcal{D})$ as a function of the diversity parameter $\lambda$, computed via the formula in Theorem 5. Each graph clearly demonstrates the behaviour predicted by Theorem 5 concerning the degrees of freedom. In particular, the degrees of freedom function $\lambda \mapsto \text{df}(\boldsymbol{F}_\lambda, \mathcal{D})$ is continuous, strictly increasing and strictly convex. Figure 5 shows the training error $R_{\text{emp}}(\boldsymbol{F}_\lambda, \mathcal{D})$ as a function of the diversity parameter $\lambda$. As predicted by Theorem 5, the function $\lambda \mapsto R_{\text{emp}}(\boldsymbol{F}_\lambda, \mathcal{D})$ is strictly decreasing.

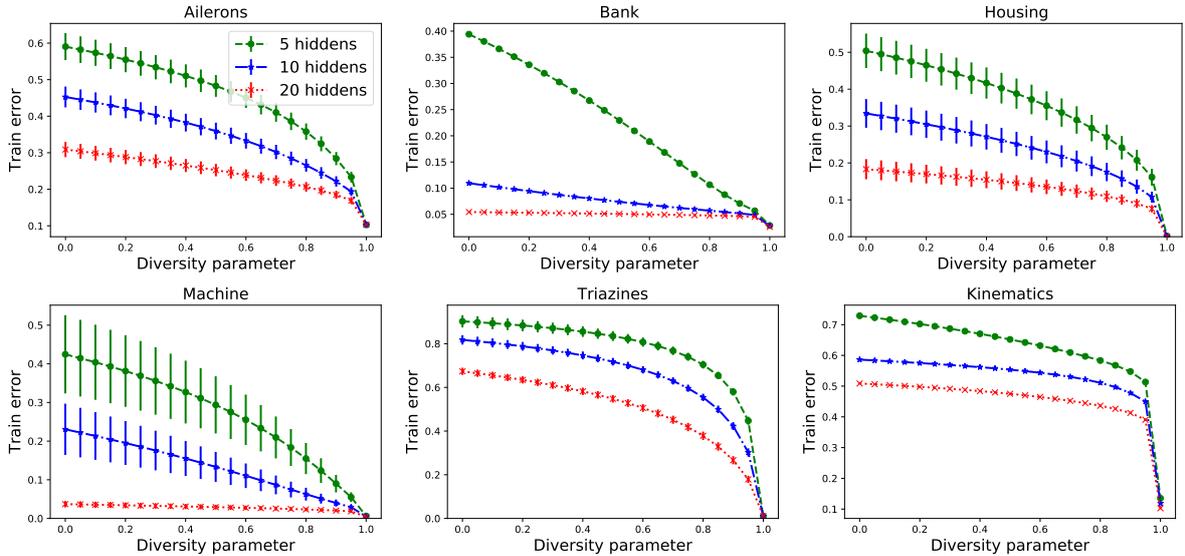

Figure 5: The training error $R_{\text{emp}}(\boldsymbol{F}_\lambda, \mathcal{D})$ as a function of the diversity parameter $\lambda$. For details see section 6.

## 7. Using the degrees of freedom formula to tune the diversity parameter

In this section we give a practical application for the explicit formula for degrees of freedom presented in Theorem 5. We shall use the degrees of freedom formula, combined with Stein's unbiased risk estimate (see Definition 2) to tune the diversity parameter. We consider a range of data sets on NCL ensembles consisting of $M = 100$ networks, each with $H = 10$ hidden nodes. As in Section 6 we use basis functions $\phi_m : \mathcal{X} \to \mathbb{R}^{H \times 1}$ formed by taking an $H$-tuple of Random Fourier Features [36], with the frequency parameter set via the method of Kwok and Tsang [29]. We compare two approaches to obtaining a proxy for the true out of sample error. First, we have the baseline method in which we perform 5-fold cross validation data within the training data. Second, we have the method based on Stein's unbiased risk estimate (SURE) in which we compute the degrees of freedom using Theorem 5. In the latter approach we use the estimate for variance obtained by $\tilde{\sigma}^2 := \left(\sum_{n=1}^{N} F_0(x_n) - y_n\right)/(N - H)$, following the approach of Ye [48, equation (20)] applied to $F_0$. In each case the diversity parameter is selected by applying the Brent minimisation routine to the corresponding criterion [1] (5-fold cross validation error or SURE). We evaluate the mean square error on the test set, with the selected diversity parameter, along with the total optimisation time, for each method. This entire procedure is repeated five times on distinct test-train splits of the data obtained via cross-validation. Note that this means in the case of the baseline method there are two nested cross-validation loops: An outer loop, which is also applied to the SURE method, to asses the efficacy of the procedure, and an inner loop in which the diversity parameter is selected based solely on the training data from the outer loop. The data sets used are described in Appendix E. Several data sets correspond to multi-output regression problems. For these data sets we opted for a simple approach in which an entirely separate NCL model was trained for each of the outputs. More nuanced approaches to multi-output regression, which take account of the dependencies between the target variables, are available [25, 42]. We leave the question of how well NCL performs in combination with these more nuanced approaches for future work. The displayed test error is obtained by taking a mean average over test errors of the different outputs.

Table 1 shows the results. Whilst both the 5-fold cross validation method and the SURE method perform similarly in terms of test error, the SURE method leads to a significant reduction in training time.

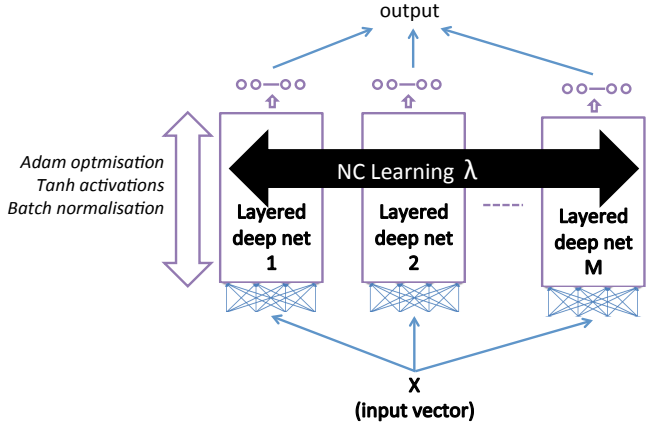

Figure 6: Illustration of our experimental setup with deep Negative Correlation Learning.

|  | Test error | | Time (seconds) | | Diversity ($\lambda$) | |
| --- | --- | --- | --- | --- | --- | --- |
| Data set | 5-fold | SURE | 5-fold | SURE | 5-fold | SURE |
| Kinematics | 0.41±0.02 | 0.41±0.02 | 29.61±3.36 | 3.75±0.41 | 0.99 | 0.99 |
| Machine | 0.76±0.83 | 0.76±0.85 | 12.82±0.42 | 3.87±0.20 | 0.40 | 0.91 |
| California | 0.32±0.03 | 0.32±0.03 | 53.11±5.05 | 4.12±0.27 | 0.99 | 0.99 |
| Wisconsin | 1.03±0.45 | 1.01±0.45 | 12.73±0.30 | 3.77±0.11 | 0.37 | 0.47 |
| Housing | 0.26±0.19 | 0.26±0.19 | 13.50±0.41 | 3.79±0.12 | 0.94 | 0.97 |
| Triazines | 0.92±0.16 | 0.89±0.10 | 12.66±0.17 | 3.95±0.22 | 0.89 | 0.75 |
| Ailerons | 0.18±0.05 | 0.18±0.05 | 27.28±2.68 | 3.66±0.29 | 0.99 | 0.99 |
| Bank | 0.05±0.00 | 0.05±0.00 | 20.73±0.81 | 3.66±0.07 | 0.99 | 0.99 |
| Elevators | 0.13±0.07 | 0.13±0.07 | 30.36±2.47 | 3.95±0.34 | 0.99 | 0.99 |
| Abalone | 0.46±0.22 | 0.46±0.22 | 20.01±1.20 | 3.71±0.13 | 0.97 | 0.99 |
| Energy | 0.08±0.01 | 0.08±0.01 | 56.47±0.27 | 14.36±0.11 | 0.99 | 0.99 |
| Andromeda | 1.05±0.59 | 1.04±0.55 | 178.50±2.53 | 46.79±0.39 | 0.40 | 0.49 |
| SCM1D | 0.41±0.05 | 0.41±0.05 | 829.73±3.94 | 124.41±0.69 | 0.99 | 0.99 |
| SCM20D | 0.57±0.05 | 0.59±0.05 | 798.43±7.41 | 128.77±0.89 | 0.95 | 0.99 |
| Water | 0.92±0.10 | 0.94±0.11 | 411.80±4.66 | 101.74±1.34 | 0.60 | 0.88 |
| OES10 | 0.93±0.46 | 0.93±0.46 | 442.90±2.96 | 119.05±0.73 | 0.81 | 0.70 |
| OES97 | 0.93±0.51 | 0.93±0.51 | 440.84±1.03 | 121.37±0.82 | 0.83 | 0.68 |
| Jura | 0.38±0.05 | 0.39±0.05 | 83.03±3.24 | 22.44±0.35 | 0.95 | 0.93 |
| Electrical | 0.69±0.47 | 0.67±0.45 | 50.54±1.10 | 14.27±0.05 | 0.49 | 0.75 |
| Slump | 0.51±0.11 | 0.52±0.11 | 81.86±1.54 | 23.14±0.29 | 0.83 | 0.72 |

Table 1: A comparison of two methods for tuning the diversity parameter: 5-fold cross validation and Stein's unbiased risk estimate with our degrees of freedom formula. Out of sample error, total training time and the mean diversity parameter for each method are displayed. See Section 7 for details.

## 8. The degrees of freedom for deep neural network ensembles

Deep neural networks play an increasingly central role in a wide variety of state of the art machine learning applications [27, 30, 41, 18]. Deep neural networks utilise multiple layers of trainable parameters to construct hierarchical representations [2, 4]. Our analytic formula for the degrees of freedom of an NCL ensemble, requires that each constituent model in the ensemble consists of a linear map composed with a fixed basis function. Thus, our formula does not apply directly to ensembles of deep neural networks. In order to investigate the relationship between de-grees of freedom and diversity for ensembles of deep neural networks we construct a Monte Carlo estimator for the degrees of freedom. By using this estimator we can extend the empirical conclusions of Section 6. In particular, we shall show that the degrees of freedom increases monotonically with the level of diversity, whilst the training error decreases monotonically with the level of diversity. Our approach is based on the estimators of Ramani [38] and Ye [48]. Ramani's approach begins with the following theorem.

**Theorem 8** (Ramani [38]). *Suppose that $g : \mathbb{R}^N \to \mathbb{R}^N$ is a twice differentiable function. Let $\boldsymbol{b}$ an independent and identically distributed random vector in $\mathbb{R}^N$ with standard Gaussian entries. Then, for all $\boldsymbol{y} = (y_n)_{n=1}^N \in \mathbb{R}^N$, we have*

$$\sum_{n=1}^{N} \frac{\partial g(\boldsymbol{y})_n}{\partial y_n} = \lim_{\epsilon \to 0} \mathbb{E}\left[\epsilon^{-1} \cdot \boldsymbol{b}^T \left(g(\boldsymbol{y} + \epsilon \cdot \boldsymbol{b}) - g(\boldsymbol{y})\right)\right].$$

Now suppose we have an estimator $\hat{\mu} : \mathcal{X} \to \mathbb{R}$. The estimator $\hat{\mu}$ implicitly depends upon the random vector of targets $\boldsymbol{y} = [y_1, \cdots, y_N] \in \mathbb{R}^N$ where $\mathcal{D} = \{(\boldsymbol{x}_n, y_n)\}_{n=1}^N$. In this section we shall consider estimators trained with different data sets and so make this dependence explicit with the superscript $\hat{\mu}^{\boldsymbol{y}}$. Given $\boldsymbol{y} \in \mathbb{R}^N$ we let $\hat{\boldsymbol{\mu}}^{\boldsymbol{y}}$ denote $[\hat{\mu}^{\boldsymbol{y}}(\boldsymbol{x}_1), \cdots, \hat{\mu}^{\boldsymbol{y}}(\boldsymbol{x}_N)]$. Theorem 8 has the following immediate corollary.

**Corollary 9.** *Suppose that $\boldsymbol{y} \mapsto \hat{\boldsymbol{\mu}}^{\boldsymbol{y}}$ is a twice differentiable function. Let $\boldsymbol{b} \in \mathbb{R}^N$ with i.i.d entries $b_n \sim \mathcal{N}(0, 1)$. Then,*

$$df(\hat{\mu}, \mathcal{D}) = \lim_{\epsilon \to 0} \mathbb{E}\left[\epsilon^{-1} \cdot \boldsymbol{b}^T \left(\hat{\boldsymbol{\mu}}^{\boldsymbol{y}+\epsilon \cdot \boldsymbol{b}} - \hat{\boldsymbol{\mu}}^{\boldsymbol{y}}\right)\right] = \lim_{\epsilon \to 0} \mathbb{E}\left[\sum_{n=1}^{N} b_n \cdot \epsilon^{-1} \cdot \left(\hat{\mu}^{\boldsymbol{y}+\epsilon \cdot \boldsymbol{b}}(x_n) - \hat{\mu}^{\boldsymbol{y}}(x_n)\right)\right].$$

Corollary 9 suggests a simple Monte-Carlo strategy for estimating $df(\hat{\mu}, \mathcal{D})$ as detailed in Algorithm 1.

**Inputs:** A data set $\mathcal{D} = \{(\boldsymbol{x}_n, y_n)\}_{n=1}^N$, an estimator $\hat{\mu}$ & $\epsilon$;
Train $\hat{\mu}^{\boldsymbol{y}}$ on the original data set $\mathcal{D}$;
Sample $\boldsymbol{b} = (b_n)_{n=1}^N \sim \mathcal{N}(\boldsymbol{0}_N, \boldsymbol{I}_{N \times N})$;
Train $\hat{\mu}^{\boldsymbol{y}+\epsilon \cdot \boldsymbol{b}}$ on the perturbed data set $\tilde{\mathcal{D}} = \{(\boldsymbol{x}_n, y_n + \epsilon \cdot b_n)\}_{n=1}^N$;
Compute $\hat{df}_{\text{Ramani}}(\hat{\mu}, \mathcal{D}) = \sum_{n=1}^{N} b_n \cdot \epsilon^{-1} \cdot \left(\hat{\mu}^{\boldsymbol{y}+\epsilon \cdot \boldsymbol{b}}(x_n) - \hat{\mu}^{\boldsymbol{y}}(x_n)\right)$;
**Output:** Degrees of freedom estimate $\hat{df}_{\text{Ramani}}(\hat{\mu}, \mathcal{D})$

**Algorithm 1:** The Monte-Carlo estimate for degrees of freedom based on Ramani [38].

This is the direct analogue of the Monte-Carlo estimate of Ramani [38] in the supervised regression setting. Recently, Gao and Jojic conducted a detailed investigation into the degrees of freedom of deep neural networks for classification, with the cross-entropy loss [16] based on a similar estimator to $\hat{df}_{\text{Ramani}}(\hat{\mu}, \mathcal{D})$. We shall apply the estimator $\hat{df}_{\text{Ramani}}(\hat{\mu}, \mathcal{D})$ to empirically study the degrees of freedom of NC ensembles of deep neural networks.

We consider deep NCL ensembles $\boldsymbol{F}_\lambda$, trained to minimise the NCL loss $L_\lambda$ (see Figure 6). In each case our architecture consisted of an ensemble of $M = 30$ networks with $H = 5$ hidden nodes per layer. The number of layers was varied between 3 and 5. Each layer was fully connected with tanh activation functions. To optimise $L_\lambda$ we employ Adam, a recent variant of stochastic gradient descent which efficiently trains deep networks by scaling weight updates by an estimate of the variance [26]. In addition we apply batch normalisation [24] to avoid internal covariate shift. Without these tools optimisation of $\boldsymbol{F}_\lambda$ was extremely slow. Following Bengio [3], the initial learning rate was selected from $\{10^{-6}, \cdots, 10^{-1}, 1\}$, and the learning rate was halved after 100 epochs without a reduction in the cost $L_\lambda$. The weights were initialised using the method of Mishkin [33].

Figure 7 shows the degrees of freedom estimate $\hat{df}_{\text{Ramani}}(\hat{\mu}, \mathcal{D})$, estimated via Algorithm 1, as a function of the diversity parameter $\lambda$. The mean and standard error are displayed, based on a five-fold cross-validation procedure. For each data set and each number of layers we observe that the degrees of freedom estimate $\hat{df}_{\text{Ramani}}(\hat{\mu}, \mathcal{D})$ is monotonically increasing as a function of the diversity parameter. This conforms to the pattern observed in the case of fixed basis functions in Figure 4. Figure 8 shows the training error $R_{\text{emp}}(\boldsymbol{F}_\lambda, \mathcal{D})$ as a function of the diversity parameter. In each case the training error is montonically decreasing as a function of the diversity parameter. Again, this matches the pattern observed in the case of fixed basis functions observed in Figure 5. Thus, whilst the assumptions of Theorem 5 do not extend to ensembles of deep neural networks our empirical results give strong evidence that the monotonicity of both the degrees of freedom and training error holds in this more general setting.

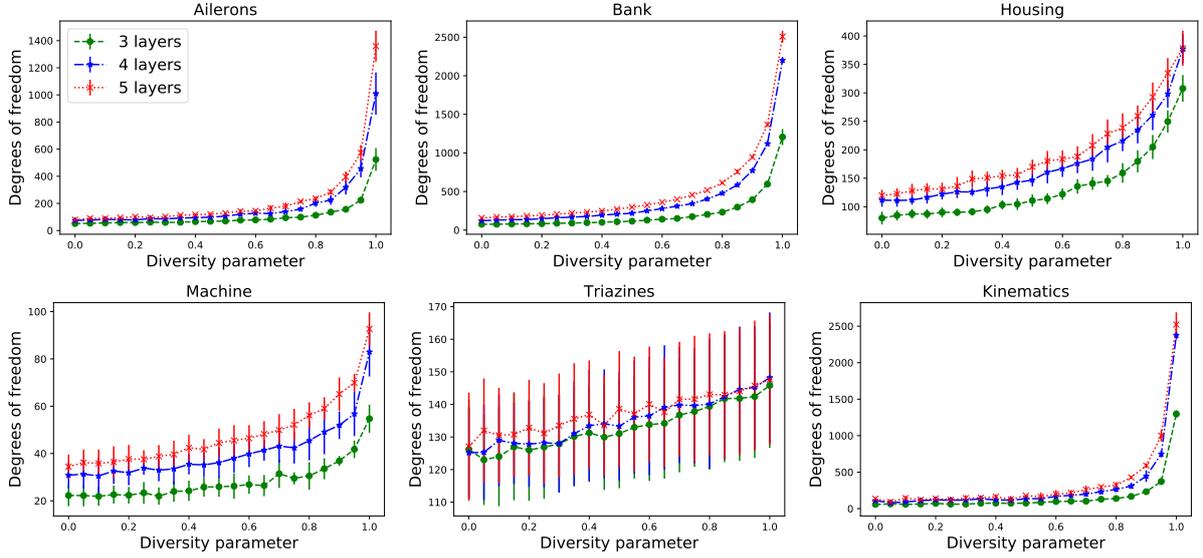

Figure 7: The degrees of freedom estimator $\hat{\mathrm{df}}_{\mathrm{Ramani}}\left(\boldsymbol{F}_\lambda, \mathcal{D}\right)$ for ensembles of deep NCL networks, as a function of the diversity parameter $\lambda$. See Section 8 for details.

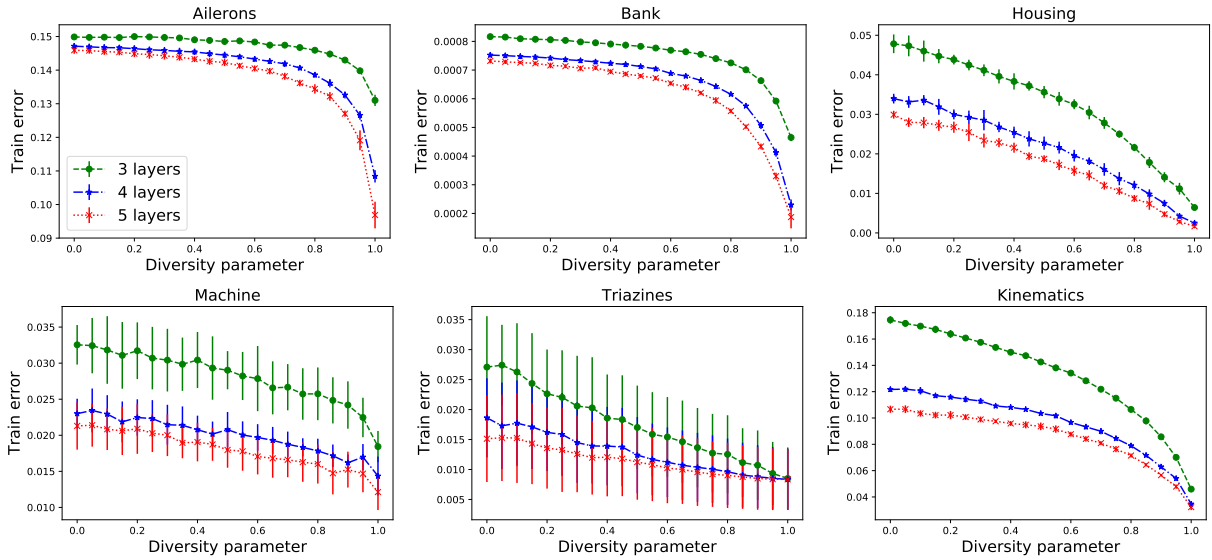

Figure 8: The training error $R_{\mathrm{emp}}\left(\boldsymbol{F}_\lambda, \mathcal{D}\right)$ for ensembles of deep NCL networks, as a function of the diversity parameter $\lambda$. See Section 8 for details.

## 9. Discussion

In this paper we have addressed several previously unanswered questions concerning Negative Correlation Learning, a highly effective method for training neural network ensembles. The core of our approach lies in a degrees of freedom analysis of NCL. We gave an explicit formula for the degrees of freedom of NCL ensembles with fixed basis functions and showed that the degrees of freedom a is monotonically increasing and convex function of the diversity parameter. We then confirmed these results empirically on a range of data sets. We demonstrated that our degrees of freedom formula may be utilised to tune the diversity parameter in a way that is both fast and effective. We then extended our empirical analysis to ensembles of deep neural networks by using a Monte-Carlo estimator for the degrees of freedom due to Ramani [38]. We also presented the surprising result that the optimal value of the diversity parameter is always strictly below one, whenever there is positive noise in the data. Finally, we presented an interesting connection to Tikhonov regularisation. Overall, we developed a deeper understanding of how the statistical behaviour of NCL ensembles depends upon the level of diversity, through a degrees of freedom analysis, showing that diversity acts as a form of *inverse regularisation*.

## 10. Acknowledgements

The authors gratefully acknowledge the support of the EPSRC for the LAMBDA project (EP/N035127/1) and the Manchester Centre for Doctoral Training in Computer Science (EP/1038099/1). We would also like to thank Konstantinos Sechidis, Nikos Nikolaou and the three anonymous reviewers for their careful feedback.

## Appendix A. Proof of Theorem 4

In this section we focus on a special class of NCL ensembles $\mathcal{F}$ in which each function $f_m$ consists of a linear map applied to fixed basis function. That is, for each $m = 1, \cdots, M$ there exists a fixed function $\phi_m : \mathcal{X} \to \mathbb{R}^{H \times 1}$ with $f_m(x) = \langle w_m, \phi_m(x) \rangle$, where $w_m \in \mathbb{R}^{H \times 1}$ is a trainable weight vector. In this situation the ensemble $F$ which minimises $L_\lambda$ averaged over the data has a simple closed form solution. We first introduce some notation. Let $Q = H \cdot M$ and let $\boldsymbol{\phi} : \mathcal{X} \to \mathbb{R}^{Q \times 1}$ be the function defined by $\boldsymbol{\phi}(x) = [\phi_1(x)^T, \cdots, \phi_M(x)^T]^T$. We let

$$\langle \boldsymbol{\phi}, \boldsymbol{\phi} \rangle_\mathcal{D} := \frac{1}{N} \sum_{n=1}^{N} \boldsymbol{\phi}(x_n) \boldsymbol{\phi}(x_n)^T \in \mathbb{R}^{Q \times Q}$$

$$\langle \boldsymbol{\phi}, \boldsymbol{y} \rangle_\mathcal{D} := \frac{1}{N} \sum_{n=1}^{N} y_n \boldsymbol{\phi}(x_n)^T \in \mathbb{R}^{Q \times 1}$$

Similarly, for each $m = 1, \cdots, M$, we define

$$\langle \phi_m, \phi_m \rangle_\mathcal{D} := \frac{1}{N} \sum_{n=1}^{N} \phi_m(x_n) \phi_m(x_n)^T \in \mathbb{R}^{H \times H}$$

$$\langle \boldsymbol{\phi}, \boldsymbol{\phi} \rangle_\mathcal{D}^{\text{diag}} := \text{diag}\left(\langle \phi_1, \phi_1 \rangle_\mathcal{D}, \cdots, \langle \phi_M, \phi_M \rangle_\mathcal{D}\right) \in \mathbb{R}^{Q \times Q}.$$

We shall assume that $\langle \phi_m, \phi_m \rangle_\mathcal{D}$ is of full rank $H$. Under these assumptions the NCL estimator has a closed form solution. Note that given a matrix $A$, $A^+$ denotes the Moore-Penrose psuedoinverse of $A$.

**Theorem 4.** *Suppose that each function $f_m$ consists of a linear map applied to a fixed basis function $\phi_m$. For each $\lambda \in [0, 1]$, the ensemble function $F_\lambda : \mathcal{X} \to \mathbb{R}$ is given by $F_\lambda(x) = \langle \beta_\lambda, \boldsymbol{\phi}(x) \rangle$ where*

$$\beta_\lambda = \left(M(1 - \lambda) \cdot \langle \boldsymbol{\phi}, \boldsymbol{\phi} \rangle_\mathcal{D}^{\text{diag}} + \lambda \langle \boldsymbol{\phi}, \boldsymbol{\phi} \rangle_\mathcal{D}\right)^+ \langle \boldsymbol{\phi}, \boldsymbol{y} \rangle_\mathcal{D}.$$

*Proof of Theorem 4.* Recall that the NCL loss is given by,

$$L_\lambda(F, x, y) = \frac{1 - \lambda}{M} \cdot \sum_{q=1}^{M} \left(f_q(x) - y\right)^2 + \lambda \cdot (F(x) - y)^2.$$

Hence, we have

$$\frac{\partial L_\lambda(F, x, y)}{\partial f_m(x)} = \frac{2}{M} \cdot \left(((1 - \lambda) \cdot f_m(x) + \lambda \cdot F(x)) - y\right).$$

Note that $f_m(x) = w_m^T \phi_m(x)$, so $\partial f_m(x)/\partial w_m = \phi_m(x)$, so

$$\frac{\partial L_\lambda(F, x, y)}{\partial w_m} = \frac{2}{M} \cdot \left(((1 - \lambda) \cdot f_m(x) + \lambda \cdot F(x)) - y\right) \phi_m(x)^T$$

$$= \frac{2}{M} \cdot \left(\left((1 - \lambda) \cdot w_m^T \phi_m(x) + \lambda \cdot \frac{1}{M} w^T \boldsymbol{\phi}(x)\right) - y\right) \phi_m(x).$$

Here, $w$ denotes $w := \left[w_1^T, \cdots, w_M^T\right]^T \in \mathbb{R}^{Q \times 1}$. Hence, to minimise the NCL loss $L_\lambda$ averaged over the data $\mathcal{D}$, for each $m$ we must have,

$$\mathbf{0}_{H \times 1} = \frac{1}{N} \sum_{n=1}^{N} \phi_m(x_n) \left(\left((1 - \lambda) \cdot \phi_m(x_n)^T w_m + \lambda \cdot \frac{1}{M} \boldsymbol{\phi}(x_n)^T w\right) - y_n\right)$$

$$= (1 - \lambda) \cdot \langle \phi_m, \phi_m \rangle_\mathcal{D} w_m + \lambda \frac{1}{M} \langle \boldsymbol{\phi}, \phi_m \rangle_\mathcal{D} w - \langle \phi_m, y \rangle_\mathcal{D}.$$

This is equivalent to

$$\mathbf{0}_{Q\times 1} = \left((1-\lambda)\cdot \langle \boldsymbol{\phi},\boldsymbol{\phi}\rangle_{\mathcal{D}}^{\text{diag}} + \lambda \frac{1}{M}\langle \boldsymbol{\phi},\boldsymbol{\phi}\rangle_{\mathcal{D}}\right)\mathbf{w} - \langle \boldsymbol{\phi},\mathbf{y}\rangle_{\mathcal{D}}.$$

Thus, at the minimum we have,

$$\mathbf{w} = M \cdot \left(M(1-\lambda)\cdot \langle \boldsymbol{\phi},\boldsymbol{\phi}\rangle_{\mathcal{D}}^{\text{diag}} + \lambda \cdot \langle \boldsymbol{\phi},\boldsymbol{\phi}\rangle_{\mathcal{D}}\right)^{+} \langle \boldsymbol{\phi},\mathbf{y}\rangle_{\mathcal{D}}.$$

Taking $\beta_\lambda = (1/M)\mathbf{w}$ proves the theorem. $\square$

**Definition 10** (Linear smoother). *A linear smoother is an estimator $\hat{\mu}$ such that there exists an $N \times N$ matrix $S(X)$, which depends upon $X = [x_1, \cdots, x_N]$, but not on $[y_1, \cdots, y_N]$ such that $[\hat{\mu}(x_1), \cdots, \hat{\mu}(x_N)] = [y_1, \cdots, y_N]S(X)$.*

For each $m \in \{1, \cdots, M\}$ let $\Phi_m := [\phi_m(x_1), \cdots, \phi_m(x_N)] \in \mathbb{R}^{H\times N}$ and let $\boldsymbol{\Phi} := \left[\Phi_1^T, \cdots, \Phi_M^T\right]^T \in \mathbb{R}^{Q\times N}$.

**Corollary 11.** *Suppose that each function $f_m$ consists of a linear map applied to a fixed basis function $\phi_m$. For each $\lambda \in [0, 1]$, the ensemble function $F_\lambda$ is a linear smoother with smoothing matrix*

$$S_\lambda(X) := \frac{1}{N} \cdot \boldsymbol{\Phi}^T \left(M(1-\lambda)\cdot \langle \boldsymbol{\phi},\boldsymbol{\phi}\rangle_{\mathcal{D}}^{\text{diag}} + \lambda \langle \boldsymbol{\phi},\boldsymbol{\phi}\rangle_{\mathcal{D}}\right)^{+} \boldsymbol{\Phi}.$$

*Proof.* Corollary 11 follows immediately from Theorem 4. $\square$

## Appendix B. Proof of Theorem 5

In order to prove Theorem 5 we require some supporting results.

**Proposition 12.** *Suppose that $\hat{\mu}$ is a linear smoother with smoothing matrix $S(X)$ and let $\sigma^2 = \mathbb{E}\left[\epsilon_n^2\right]$, the noise variance. Define $\boldsymbol{\mu}, \mathbf{y} \in \mathbb{R}^{1\times N}$ by $\boldsymbol{\mu} := [\mu(x_1), \cdots, \mu(x_N)]$ and $\mathbf{y} := [y_1, \cdots, y_N]$. Then, we have*

1. *$df(\hat{\mu}, \mathcal{D}) = \text{trace}(S(X))$,*

2. *$R_{emp}(\hat{\mu}, \mathcal{D}) = \frac{1}{N}\mathbf{y}(I_{N\times N} - S(X))^2 \mathbf{y}^T$.*

*Proof.* The proposition follows immediately from the definitions. $\square$

**Lemma 13.** *The matrix $\boldsymbol{P} = \left(\langle \boldsymbol{\phi},\boldsymbol{\phi}\rangle_{\mathcal{D}}^{diag}\right)^{-1/2}\langle \boldsymbol{\phi},\boldsymbol{\phi}\rangle_{\mathcal{D}}\left(\langle \boldsymbol{\phi},\boldsymbol{\phi}\rangle_{\mathcal{D}}^{diag}\right)^{-1/2}$ has spectral norm $\|\boldsymbol{P}\|_{spectral} \leq M$.*

*Proof.* For each $m = 1, \cdots, M$ we take a singular value decomposition

$$\Phi_m = U_m[\Sigma_m, \mathbf{0}_{H\times(N-H)}]V_m^T,$$

where $U_m$ and $V_m$ are orthonormal and $\Sigma_m$ is diagonal. By construction we have,

$$\langle \boldsymbol{\phi},\boldsymbol{\phi}\rangle_{\mathcal{D}} = \frac{1}{N}\cdot \boldsymbol{\Phi}\boldsymbol{\Phi}^T = \frac{1}{N}\cdot \begin{bmatrix} \Phi_1\Phi_1^T & \Phi_1\Phi_2^T & \cdots & \Phi_1\Phi_M^T \\ \Phi_2\Phi_1^T & \Phi_2\Phi_2^T & \ddots & \vdots \\ \vdots & \ddots & \ddots & \vdots \\ \Phi_M\Phi_1^T & \cdots & \cdots & \Phi_M\Phi_M^T \end{bmatrix}.$$

Similarly,

$$\langle \boldsymbol{\phi},\boldsymbol{\phi}\rangle_{\mathcal{D}}^{\text{diag}} = \frac{1}{N}\cdot \begin{bmatrix} \Phi_1\Phi_1^T & \mathbf{0}_{H\times H} & \cdots & \mathbf{0}_{H\times H} \\ \mathbf{0}_{H\times H} & \Phi_2\Phi_2^T & \ddots & \vdots \\ \vdots & \ddots & \ddots & \vdots \\ \mathbf{0}_{H\times H} & \cdots & \cdots & \Phi_M\Phi_M^T \end{bmatrix}.$$

Hence, we have

$$P = \begin{bmatrix} P_{1,1} & P_{1,2} & \cdots & P_{1,M} \\ P_{2,1} & P_{2,2} & \ddots & \vdots \\ \vdots & \ddots & \ddots & \vdots \\ P_{M,1} & \cdots & \cdots & P_{M,M} \end{bmatrix},$$

where $P_{l,m} = \left(\Phi_l \Phi_l^T\right)^{-1/2} \Phi_l \Phi_m^T \left(\Phi_m \Phi_m^T\right)^{-1/2}$.

Given $m \in \{1, \cdots, M\}$ we have $\Phi_m \Phi_m^T = U_m \Sigma_m^2 U_m^T$, so $\left(\Phi_m \Phi_m^T\right)^{-1/2} = U_m \Sigma_m^{-1} U_m^T$. Hence, for each $m$,

$$\left(\Phi_m \Phi_m^T\right)^{-1/2} \Phi_m = \left(U_m \Sigma_m^{-1} U_m^T\right)\left(U_m [\Sigma_m, \mathbf{0}_{H \times ((H-1)M)}] V_m^T\right)$$
$$= U_m [I_{H \times H}, \mathbf{0}_{H \times (N-H)}] V_m^T.$$

Since $\left(\Phi_m \Phi_m^T\right)^{-1/2} \Phi_m$ is the composition of a projection and an orthogonal matrix we have

$$1 = \sup\left\{\left(\Phi_m \Phi_m^T\right)^{-1/2} \Phi_m v : v \in \mathbb{R}^N, \|v\|_2 \leq 1\right\}$$
$$= \sup\left\{\Phi_m^T \left(\Phi_m \Phi_m^T\right)^{-1/2} u : u \in \mathbb{R}^H, \|u\|_2 \leq 1\right\}.$$

Thus, given any pair $l, m \in \{1, \cdots, M\}$, we have $\|P_{l,m}\|_{\text{spectral}} \leq 1$. Note also that $P_{m,m} = I_{H \times H}$ for each $m$. In order to bound $\|P\|_{\text{spectral}}$ we first note that $P$ is a real symmetric matrix, so the spectral norm is equal to the modulus of the greatest eigenvalue. Take an eigenvector of $P$, $v = [v_1^T, \cdots, v_M^T]^T \in \mathbb{R}^{HM}$ with each $v_m \in \mathbb{R}^M$, so $Pv = \rho v$. Choose $m$ so that $\|v_m\|_2$ is maximal. We have,

$$\rho v_m = \sum_{q=1}^{M} P_{m,q} v_q.$$

Hence, using the fact that $\|P_{m,q}\|_{\text{spectral}} \leq 1$ we have

$$\rho \|v_m\| \leq \sum_{q=1}^{M} \|P_{m,q} v_q\| \leq \sum_{q=1}^{M} \|v_q\| \leq M \cdot \|v_m\|.$$

This holds for all eigenvalues $\rho$. Hence $\|P\|_{\text{spectral}} \leq M$. □

**Lemma 14.** *The matrix $P = \left(\langle \boldsymbol{\phi}, \boldsymbol{\phi} \rangle_{\mathcal{D}}^{diag}\right)^{-1/2} \langle \boldsymbol{\phi}, \boldsymbol{\phi} \rangle_{\mathcal{D}} \left(\langle \boldsymbol{\phi}, \boldsymbol{\phi} \rangle_{\mathcal{D}}^{diag}\right)^{-1/2}$ has $\text{trace}(P) = HM$ and $\text{rank}(P) = \text{rank}\left(\langle \boldsymbol{\phi}, \boldsymbol{\phi} \rangle_{\mathcal{D}}\right)$.*

*Proof.* We begin by showing that $\text{trace}(P) = HM$. As noted in the proof of Lemma 13 we have

$$P = \begin{bmatrix} P_{1,1} & P_{1,2} & \cdots & P_{1,M} \\ P_{2,1} & P_{2,2} & \ddots & \vdots \\ \vdots & \ddots & \ddots & \vdots \\ P_{M,1} & \cdots & \cdots & P_{M,M} \end{bmatrix},$$

where $P_{l,m} = \left(\Phi_l \Phi_l^T\right)^{-1/2} \Phi_l \Phi_m^T \left(\Phi_m \Phi_m^T\right)^{-1/2}$. In particular, for each $m \in \{1, \cdots, M\}$, $P_{m,m} = I_{H \times H}$, where we have used the fact that each $\langle \phi_m, \phi_m \rangle_{\mathcal{D}} = (1/N) \Phi_l \Phi_l^T$ is of full rank $H$. Thus, $P$ consists of 1s along the diagonal, so $\text{trace}(P) = HM$.

Secondly, we show that $\text{rank}(P) = \text{rank}(\langle \boldsymbol{\phi}, \boldsymbol{\phi} \rangle_{\mathcal{D}})$. For each $m \in \{1, \cdots, M\}$, $\langle \phi_m, \phi_m \rangle_{\mathcal{D}}$ is of full rank $H$. Hence, $\langle \boldsymbol{\phi}, \boldsymbol{\phi} \rangle_{\mathcal{D}}^{\text{diag}} = \text{diag}\left(\langle \phi_1, \phi_1 \rangle_{\mathcal{D}}, \cdots, \langle \phi_M, \phi_M \rangle_{\mathcal{D}}\right)$ is of full rank $HM$. Thus, by the definition of $P$ we see that $\text{rank}(P) = \text{rank}\left(\langle \boldsymbol{\phi}, \boldsymbol{\phi} \rangle_{\mathcal{D}}\right)$. □

Given a matrix $M$ we let $\text{spec}(M)$ denote its spectrum ie. its eigenvalues.

**Proposition 15.** *For each $\lambda \in [0, 1)$ the spectrum of $S_\lambda(X)$ is given by*

$$\text{spec}(S_\lambda(X)) = \left\{ \frac{\rho}{M \cdot (1 - \lambda) + \lambda \cdot \rho} : \rho \in \text{spec}(P) \right\}.$$

*Proof.* We have

$$\begin{aligned}
S_\lambda(X) &:= \frac{1}{N} \cdot \mathbf{\Phi}^T \left( M(1 - \lambda) \cdot \langle \boldsymbol{\phi}, \boldsymbol{\phi} \rangle_{\mathcal{D}}^{\text{diag}} + \lambda \langle \boldsymbol{\phi}, \boldsymbol{\phi} \rangle_{\mathcal{D}} \right)^+ \mathbf{\Phi} \\
&= \frac{1}{N} \cdot \left( \left( \langle \boldsymbol{\phi}, \boldsymbol{\phi} \rangle_{\mathcal{D}}^{\text{diag}} \right)^{-1/2} \mathbf{\Phi} \right)^T \left( M(1 - \lambda) \cdot I_{Q \times Q} + \lambda \left( \langle \boldsymbol{\phi}, \boldsymbol{\phi} \rangle_{\mathcal{D}}^{\text{diag}} \right)^{-1/2} \langle \boldsymbol{\phi}, \boldsymbol{\phi} \rangle_{\mathcal{D}} \left( \langle \boldsymbol{\phi}, \boldsymbol{\phi} \rangle_{\mathcal{D}}^{\text{diag}} \right)^{-1/2} \right)^+ \left( \left( \langle \boldsymbol{\phi}, \boldsymbol{\phi} \rangle_{\mathcal{D}}^{\text{diag}} \right)^{-1/2} \mathbf{\Phi} \right) \\
&= \frac{1}{N} \cdot \tilde{\mathbf{\Phi}}^T \left( M(1 - \lambda) \cdot I_{Q \times Q} + \lambda P \right)^+ \tilde{\mathbf{\Phi}},
\end{aligned}$$

where $\tilde{\mathbf{\Phi}} = \left( \langle \boldsymbol{\phi}, \boldsymbol{\phi} \rangle_{\mathcal{D}}^{\text{diag}} \right)^{-1/2} \mathbf{\Phi}$. Note that $P = \left( (1/\sqrt{N}) \cdot \tilde{\mathbf{\Phi}} \right) \left( (1/\sqrt{N}) \cdot \tilde{\mathbf{\Phi}} \right)^T$. Now take a singular value decomposition $(1/\sqrt{N}) \cdot \tilde{\mathbf{\Phi}} = U[\mathbf{\Sigma}, \mathbf{0}_{(N-Q) \times H}] V^T$. It follows from the above that $P = U \mathbf{\Sigma}^2 U^T$. Hence,

$$\begin{aligned}
S_\lambda(X) &= \frac{1}{N} \cdot \tilde{\mathbf{\Phi}}^T \left( M(1 - \lambda) \cdot I_{Q \times Q} + \lambda P \right)^+ \tilde{\mathbf{\Phi}} \\
&= V[\mathbf{\Sigma}, \mathbf{0}_{(N-Q) \times H}]^T U^T \left( M(1 - \lambda) \cdot I_{Q \times Q} + \lambda U \mathbf{\Sigma}^2 U^T \right)^{-1} \left( U[\mathbf{\Sigma}, \mathbf{0}_{(N-Q) \times H}] V^T \right) \\
&= V[I_{Q \times Q}, \mathbf{0}_{(N-H) \times H}]^T \mathbf{\Sigma}^2 \left( M(1 - \lambda) \cdot I_{Q \times Q} + \lambda \mathbf{\Sigma}^2 \right)^{-1} \left( [I_{Q \times Q}, \mathbf{0}_{(N-Q) \times Q}] V^T \right).
\end{aligned}$$

Hence, the non-zero eigen values of $S_\lambda(X)$ are all of the form $s^2 / \left( M \cdot (1 - \lambda) + \lambda \cdot s^2 \right)$, where $s$ is a diagonal element of $\mathbf{\Sigma}$. Moreover $P = U \mathbf{\Sigma}^2 U^T$, so the square diagonal elements of $\mathbf{\Sigma}$ are precisely the eigenvalues of $P$ ie. $\rho = s^2$. Hence, the proposition holds. $\square$

**Theorem 5.** *Suppose that each function $f_m$ is a linear map of a fixed basis function. Given any $\lambda \in [0, 1]$, the degrees of freedom for $F_\lambda$ is given by*

$$\text{df}(F_\lambda, \mathcal{D}) = \text{trace}\left( \langle \boldsymbol{\phi}, \boldsymbol{\phi} \rangle_{\mathcal{D}} \left( M(1 - \lambda) \cdot \langle \boldsymbol{\phi}, \boldsymbol{\phi} \rangle_{\mathcal{D}}^{\text{diag}} + \lambda \langle \boldsymbol{\phi}, \boldsymbol{\phi} \rangle_{\mathcal{D}} \right)^+ \right).$$

*In particular, we have the following behaviour*

- *The function $\lambda \mapsto \text{df}(F_\lambda, \mathcal{D})$ is continuous, increasing and convex;*
- *The function $\lambda \mapsto R_{\text{emp}}(F_\lambda, \mathcal{D})$ is continuous and decreasing.*

*Moreover, if $H < \text{rank}(\langle \boldsymbol{\phi}, \boldsymbol{\phi} \rangle_{\mathcal{D}})$ then,*

- *The function $\lambda \mapsto \text{df}(F_\lambda, \mathcal{D})$ is strictly increasing and strictly convex;*
- *The function $\lambda \mapsto R_{\text{emp}}(F_\lambda, \mathcal{D})$ is strictly decreasing.*

*Proof of Theorem 5.* The proof of the equation for $\text{df}(F_\lambda, \mathcal{D})$ follows immediately from Corollary 11 combined with Proposition 12, along with the fact that the trace is invariant to cyclical permutations. Moreover, letting $\{\rho_q\}_{q=1}^Q$ denote the eigenvalues of $P$. By Proposition 15 the eigenvalues of $S_\lambda(X)$ are given by $\{\rho_q / (M(1 - \lambda) + \rho_q)\}_{q=1}^Q$. Hence, we have

$$\text{df}(F_\lambda, \mathcal{D}) = \text{trace}(S_\lambda(X)) = \sum_{q=1}^Q \frac{\rho_q}{M(1 - \lambda) + \lambda \cdot \rho_q}.$$

The continuity of $\mathrm{df}(F_\lambda, \mathcal{D})$ is immediate. Taking first and second derivatives with respect to $\lambda$ we have

$$\frac{\partial \mathrm{df}(F_\lambda, \mathcal{D})}{\partial \lambda} = \sum_{q=1}^{Q} \frac{\rho_q (M - \rho_q)}{\left(M(1-\lambda) + \lambda \cdot \rho_q\right)^2}$$

$$\frac{\partial^2 \mathrm{df}(F_\lambda, \mathcal{D})}{\partial \lambda^2} = \sum_{q=1}^{Q} \frac{\rho_q (M - \rho_q)^2}{\left(M(1-\lambda) + \lambda \cdot \rho_q\right)^3}.$$

By Lemma $\rho_q \leq M$ for each $q$, so the first and second derivatives are positives ie. $\lambda \mapsto \mathrm{df}(F_\lambda, \mathcal{D})$ is monotonically increasing and convex. Moreover, by Lemma 14, we have $\sum_{q=1}^{Q} \rho_q = \mathrm{trace}(\mathbf{P}) = H \cdot M$. Moreover, if $H < \mathrm{rank}(\langle \phi, \phi \rangle_\mathcal{D})$ then the number of posiive eigen values of $\rho_q$ must exceed $H$, so there must exist at least one $q$ with $\rho_q \in (0, M)$. Hence the first and second derivatives of $\lambda \mapsto \mathrm{df}(F_\lambda, \mathcal{D})$ must be *strictly* positive, which implies that the function is strictly increasing and strictly convex.

To prove the results regarding $R_{\mathrm{emp}}(\hat{\mu}, \mathcal{D})$ we apply Lemma 12 and write

$$R_{\mathrm{emp}}(\hat{\mu}, \mathcal{D}) = \frac{1}{N} \mathbf{y} (\mathbf{I}_{N \times N} - \mathbf{S}(X))^2 \mathbf{y}^T = (1/N) \cdot \left( \|\mathbf{y}\|_2^2 - 2\left(\mathbf{y} \mathbf{S}_\lambda(X) \mathbf{y}^T\right) + \left(\mathbf{y} \mathbf{S}_\lambda(X)^2 \mathbf{y}^T\right) \right).$$

Applying Proposition 15 we let $\{v_q\}_{q=1}^{Q}$ denote the eigenvectors corresponding to the eigenvalues $\{\rho_q / \left(M(1-\lambda) + \rho_q\right)\}_{q=1}^{Q}$ of $\mathbf{S}_\lambda(X)$. We may write $\mathbf{y}$ as $(1/\sqrt{N}) \cdot \mathbf{y} = \sum_{q=1}^{Q} \kappa_q v_q + v_{\mathrm{orth}}$ where $v_{\mathrm{orth}}$ is orthogonal to $\{v_q\}_{q=1}^{Q}$. Hence,

$$R_{\mathrm{emp}}(\hat{\mu}, \mathcal{D}) = (1/N) \cdot \left( \|\mathbf{y}\|_2^2 - 2\left(\mathbf{y} \mathbf{S}_\lambda(X) \mathbf{y}^T\right) + \left(\mathbf{y} \mathbf{S}_\lambda(X)^2 \mathbf{y}^T\right) \right)$$

$$= (1/N) \cdot \|\mathbf{y}\|_2^2 + \sum_{q=1}^{Q} \kappa_q^2 \left( \frac{\rho_q^2}{\left(M(1-\lambda) + \lambda \cdot \rho_q\right)^2} - \frac{2\rho_q}{M(1-\lambda) + \lambda \cdot \rho_q} \right).$$

Hence, we have

$$\frac{\partial R_{\mathrm{emp}}(\hat{\mu}, \mathcal{D})}{\partial \lambda} = \sum_{q=1}^{Q} 2\kappa_q^2 \rho_q (M - \rho_q) \left( \frac{\rho_q}{\left(M(1-\lambda) + \lambda \cdot \rho_q\right)^3} - \frac{1}{\left(M(1-\lambda) + \lambda \cdot \rho_q\right)^2} \right)$$

$$= - \sum_{q=1}^{Q} \frac{2\kappa_q^2 \rho_q (M - \rho_q)^2 (1-\lambda)}{\left(M(1-\lambda) + \lambda \cdot \rho_q\right)^3}.$$

Thus, $R_{\mathrm{emp}}(\hat{\mu}, \mathcal{D})$ is monotonically decreasing with $\lambda$. $\square$

### Appendix C. Proof of Theorem 6

**Theorem 6.** *Suppose that each function $f_m$ is a linear map of a fixed basis function and let $\sigma^2 = \mathbb{E}\left[\epsilon_n^2\right]$, the noise variance. Then $1 \in \mathrm{argmin}\{\mathbb{E}[R_{true}(F_\lambda)] : \lambda \in [0,1]\}$ if and only if $\sigma = 0$.*

*Proof.* By Theorem 3

$$\mathbb{E}[R_{\mathrm{true}}(\hat{\mu})] = \mathbb{E}\left[ R_{\mathrm{emp}}(\hat{\mu}, \mathcal{D}) + \sigma^2 \cdot \left( \frac{2}{N} \cdot \mathrm{df}(\hat{\mu}, \mathcal{D}) - 1 \right) \right].$$

Thus, we have

$$\frac{\partial \mathbb{E}[R_{\mathrm{true}}(\hat{\mu})]}{\partial \lambda} = \mathbb{E}\left[ \frac{\partial R_{\mathrm{emp}}(\hat{\mu}, \mathcal{D})}{\partial \lambda} \right] + \frac{2\sigma^2}{N} \cdot \frac{\partial \mathrm{df}(\hat{\mu}, \mathcal{D})}{\partial \lambda}.$$

In the proof of Theorem 5 we saw that

$$\frac{\partial R_{\text{emp}}(\hat{\mu}, \mathcal{D})}{\partial \lambda} = -\sum_{q=1}^{Q} \frac{2\kappa_q^2 \rho_q (M - \rho_q)^2 (1-\lambda)}{\left(M(1-\lambda) + \lambda \cdot \rho_q\right)^3}$$

Thus, for all data sets $\mathcal{D}$, at $\lambda = 1$ we have $\partial R_{\text{emp}}(\hat{\mu}, \mathcal{D})/\partial \lambda = 0$, so at $\lambda = 1$ we have $\mathbb{E}\left[\partial R_{\text{emp}}(\hat{\mu}, \mathcal{D})/\partial \lambda\right] = 0$. On the other hand, we also saw in the proof of Theorem 5 that

$$\frac{\partial \text{df}(F_\lambda, \mathcal{D})}{\partial \lambda} = \sum_{q=1}^{Q} \frac{\rho_q (M - \rho_q)}{\left(M(1-\lambda) + \lambda \cdot \rho_q\right)^2}.$$

Thus, at $\lambda = 1$ we have

$$\frac{\partial \mathbb{E}\left[R_{\text{true}}(\hat{\mu})\right]}{\partial \lambda}\bigg|_{\lambda=1} = \frac{2\sigma^2}{N} \cdot \sum_{q=1}^{Q} \rho_q^{-1}.$$

Thus, whenever $\sigma > 0$, at $\lambda = 1$ the derivative is positive, so $\mathbb{E}\left[R_{\text{true}}(\hat{\mu})\right]$ does not attain its minimum at $\lambda = 1$. $\square$

### Appendix D. Proof of Theorem 5

**Theorem 7.** *Suppose that we have a NCL ensemble $F_\lambda = (1/M) \cdot \sum_{m=1}^{M} f_m$ where each function $f_m$ is a linear map of fixed basis function $\phi_m$. Let $G_\gamma^{Tik}$ be the function obtained by Tikhonov regularisation with $G_\gamma^{Tik} = (1/M) \cdot \sum_{m=1}^{M} g_m$, where each map $g_m$ is a linear map applied to a fixed basis function $\psi_m$. Suppose that for each $m$, $\psi_m(x) = \langle \phi_m, \phi_m \rangle_{\mathcal{D}}^{-1/2} \cdot \phi_m(x)$. Then for all $\lambda \in [0, 1]$, with $\gamma = (1-\lambda)/(M^2 \cdot \lambda)$ we have $G_\gamma^{Tik} = \lambda \cdot F_\lambda$.*

*Proof.* First write $\psi(x) = (1/M) \cdot \left[\psi_1(x)^T, \cdots, \psi_M(x)^T\right]^T$ and let $w = \left[w_1^T, \cdots, w_m^T\right]^T$. Hence, we may write $G_\gamma^{Tik}(x) = w^T \psi(x)$. Moreover, we may write

$$L_\gamma^{\text{Tik}}(\mathcal{G}, x, y) = \left(w^T \psi(x) - y\right)^2 + \gamma \cdot \|w\|_2^2.$$

Thus, minimising $L_\gamma^{\text{Tik}}(\mathcal{G}, x, y)$ corresponds to ridge regression on the features $\psi(x)$. Hence, if we define $\psi(\mathcal{D}) = [\psi(x_1), \cdots, \psi(x_N)]$, then we can minimise $\frac{1}{N} \sum_{n=1}^{N} L_\gamma^{\text{Tik}}(\mathcal{G}, x_n, y_n)$ by taking

$$w = \left(\gamma I_{Q \times Q} + (1/N) \cdot \psi(\mathcal{D}) \psi(\mathcal{D})^T\right)^+ \left((1/N) \cdot \psi(\mathcal{D}) y^T\right).$$

Hence, we have

$$G_\gamma^{\text{Tik}}(x) = \left((1/N) \cdot y \psi(\mathcal{D})^T\right)\left(\gamma I_{Q \times Q} + (1/N) \cdot \psi(\mathcal{D}) \psi(\mathcal{D})^T\right)^+ \psi(x).$$

Now by the definition of $\psi_m$ and $\psi$, we may write $\psi(x) = (1/M) \cdot \left(\langle \phi, \phi \rangle_{\mathcal{D}}^{\text{diag}}\right)^{-1/2} \phi(x)$. Hence, $(1/N) \cdot y \psi(\mathcal{D})^T = (1/M) \cdot \langle \phi, y \rangle_{\mathcal{D}}^T \left(\langle \phi, \phi \rangle_{\mathcal{D}}^{\text{diag}}\right)^{-1/2}$ and $(1/N) \cdot \psi(\mathcal{D}) \psi(\mathcal{D})^T = (1/M^2) \cdot \left(\langle \phi, \phi \rangle_{\mathcal{D}}^{\text{diag}}\right)^{-1/2} \cdot \langle \phi, \phi \rangle_{\mathcal{D}} \left(\langle \phi, \phi \rangle_{\mathcal{D}}^{\text{diag}}\right)^{-1/2}$. Substituting these formulas into the above expression for $G_\gamma^{\text{Tik}}(x)$ and rearranging we have

$$G_\gamma^{\text{Tik}}(x) = \langle \phi, y \rangle_{\mathcal{D}}^T \left((\gamma \cdot M^2) \cdot \langle \phi, \phi \rangle_{\mathcal{D}}^{\text{diag}} + \langle \phi, \phi \rangle_{\mathcal{D}}\right)^+ \phi(x).$$

Hence, by Theorem 4 if we take $\gamma = (1-\lambda)/(M^2 \cdot \lambda)$ then $G_\gamma^{\text{Tik}}(x) = \lambda \cdot F_\lambda(x)$. This completes the proof of the Theorem. $\square$

## Appendix E. Data sets

In this section we give a brief description of all the data sets used in our experiments. The features and target variables of each data set are preprocessed by subtracting the mean and dividing by the standard deviation. Table E.2 gives the dimensions of our data sets.

**Kinematics**

This data set corresponds to a realistic simulation of the forward dynamics of a robot arm. The task is to predict the distance of the end-effector from a target based on a collection of continuous features describing the robot arm such as joint positions and twist angles. There are eight thousand one hundred and ninety two examples and eight features. The data set may be found at [45].

**Machine**

The task in this data set is to predict relative CPU performance based upon features of the CPU: machine cycle time, minimum main memory, maximum main memory, cache memory, minimum channels and maximum channels. The data set is originally from the UCI Machine Learning Repository [31]. We use the version from [45]. There are two hundred and nine examples and six features.

**California**

The task in this data set is to predict house price based on features such as median income of occupants, house age, number of rooms, number of bedrooms etc. The data set consists of twenty thousand, four hundred and sixty examples and eight features. It may be found at [45] and is originally from [34].

**Wisconsin**

For this regression data set the task is to predict the time until recurrence of breast cancer based on features of the tumor. The data set consists of one hundred and ninety four examples and thirty two features and may be found at [45].

**Housing**

The task in this data set is to predict house values in various areas of Boston based upon several relevant features such as local crime rate and an index of accessibility etc. There are five hundred and six examples with thirteen features. For more details see [45].

**Triazines**

In this data set the task is to predict the activity of triazenes based on a large number of descriptive structural attributes. There are one hundred and eighty six examples and sixty features. For more details see [45].

**Ailerons**

In this data set the task is to predict the control action on the ailerons of an F16 aircraft data set required to achieve a certain status. There are thirteen thousand, seven hundred and fity examples and forty features. For more details see [45].

**Bank**

This data set is based upon a simulation of customer's behavior at banks. The task is to predict the proportion of customers that are turned away from the bank because all the open tellers have full queues. There are eight thousand, one hundred and ninety two examples and nine features. The full data set may be found [45, bank-8fm].

**Elevators**

The 'Elevators' data set is closely related to the 'Airelons' and also concerns the control of an F16 aircraft. However, in this case the task is to predict the required position of the aircraft's elevators based upon a different set of features. There are sixteen thousand, five hundred and fifty nine examples and eighteen features. The data set is available at [45].

**Abalone**

In this data set the task is to predict the age of abalone based upon a set of accessible physical features. There are four thousand, one hundred and seventy seven examples ane eight features. The data set is available at [45].

**Energy**

The task in this data set is to estimate the heating and cooling load of a building based upon building parameters such as glazing area, roof area, and overall height etc. The data set consists of seven hundred and sixty eight examples, eight features and two output variables. For further details see the original paper [46] and [42]. The data set may be downloaded from [20, Enb].

**Andromeda**

In this data set the task is to predict six water quality variables (temperature, pH, conductivity, salinity, oxygen, turbidity) for six days hence, based on the value of these variables over the previous five days. For further details see [22] and [42]. The data set consists of forty nine examples, thirty features and six output variables. For further details and the data set see [20, Andro].

**Water quality**

In the water quality data set (Water) the task is to predict the representation of various plant and animal species in Slovenian rivers based upon several physical and chemical water quality parameters. For details see [13] and [42]. The data set consists of one thousand and sixty examples, sixteen features and fourteen target variables. For further details and the data set see [20, Wq].

**Occupational Employment Survey**

The Occupational Employment Survey data sets (OES97 & OES10) are based upon survey data from the 1997 and 2010 Occupational Employment Surveys, respectively, compiled by the US Bureau of Labor Statistics. OES97 and OES10 are multi-target regression problems in which the task is to estimate the number of full-time employees across many employment types for a given city based upon the estimated number of full-time equivalent employees in a wide range of different other employment types for that city. Both data sets were first used for the purpose of multi-target regression in [42]. OES97 consists of three hundred and sixty four examples, two hundred and sixty three features and sixteen target variables. OES10 consists of four hundred and three examples, two hundred and ninety eight features and sixteen target variables. Note that missing values in both the input and the target variables were replaced by sample means for these results. For further details and the data set see [42], [20, Oes97 & Oes10].

**Supply Chain Management**

The Supply Chain Management data sets (SCM1D & SCM20D) are derived from the Trading Agent Competition in Supply Chain Management (TAC SCM) tournament from 2010 [21]. The task is to predict the future price of various products from previous prices. In SCM1D one must predict the mean average prices over the next day. SCM1D consists of nine thousand, eight hundred and three examples, two hundred and eighty features and sixteen target variables. In SCM20D one must predict the mean average prices over the next twenty days. SCM20D has eight thousand, nine hundred and sixty six examples, sixty one features and sixteen target variables. For further details see [21]. The data set is available at [20, Scm1d & Scm20d].

**Jura**

The data set is based upon measurements of concentrations of heavy metals recorded at various locations in the topsoil of a region of the Swiss Jura [19]. The task is to predict the concentration of cadmium and copper (concentration of these metals is relatively difficult to measure) based on the concentration of other metals and the land use. This data set contains three hundred and fifty nine examples, fifteen features and three target variables. For more details see [18, 42]. The data set may be downloaded from [20].

**Electrical**

The Electrical Discharge Machining dataset (Electrical) gives a multivariate regression problem in which the task is to predict the behaviour of a human operator that controls the values of two variables [25]. In this data set there are one hundred and fifty four examples, sixteen features and two output variables. The data set may be downloaded from [20, Edm].

**Slump**

The Slump dataset is multi-target regression problem where the task is to determine properties of concrete (slump, flow and compressive strength) using features of the concrete's compoisition [49]. In this data set there are one hundred and three examples, seven features and three output variables. For further details see [42, 49]. The data set may be downloaded from [20].

| Data set | Number of examples | Number of features | Number of outputs |
| --- | --- | --- | --- |
| Kinematics | 8192 | 8 | 1 |
| Machine | 209 | 6 | 1 |
| California | 20460 | 8 | 1 |
| Wisconsin | 194 | 32 | 1 |
| Housing | 506 | 13 | 1 |
| Triazines | 186 | 60 | 1 |
| Ailerons | 13750 | 40 | 1 |
| Bank | 8192 | 9 | 1 |
| Elevators | 16559 | 18 | 1 |
| Abalone | 4177 | 8 | 1 |
| SCM1d | 9803 | 280 | 16 |
| SCM20d | 8966 | 61 | 16 |
| Energy | 768 | 8 | 2 |
| Andromeda | 49 | 8 | 2 |
| Water | 1060 | 16 | 2 |
| OES10 | 403 | 298 | 16 |
| OES97 | 334 | 263 | 16 |
| Jura | 359 | 15 | 3 |
| Electrical | 154 | 16 | 2 |
| Slump | 103 | 7 | 3 |

Table E.2: Contains the dimensions of the regression data sets used in experiments. For more details on the data sets see Appendix E.